\documentclass[runningheads]{llncs}
\usepackage{graphicx}
\usepackage{subfigure}

\usepackage{tikz}
\usepackage{comment}
\usepackage{amsmath,amssymb} 
\usepackage{color}
\usepackage{amsmath}
\usepackage{booktabs}
\usepackage{multirow}
\usepackage{adjustbox}
\usepackage{afterpage}

\usepackage{xr}
\makeatletter
\newcommand*{\addFileDependency}[1]{
  \typeout{(#1)}
  \@addtofilelist{#1}
  \IfFileExists{#1}{}{\typeout{No file #1.}}
}
\makeatother

\newcommand*{\myexternaldocument}[1]{
    \externaldocument{#1}
    \addFileDependency{#1.tex}
    \addFileDependency{#1.aux}
}

\myexternaldocument{supplementary}
\usepackage[pagebackref,breaklinks,colorlinks]{hyperref}

\usepackage{cleveref}
\usepackage[flushleft]{threeparttable}
\Crefname{figure}{Fig.}{Figs.}
\Crefname{table}{Tab.}{Tabs.}

\newcommand{\etal}{\textit{et al. }}

\newcommand{\eg}{\textit{e}.\textit{g}., }

\usepackage[accsupp]{axessibility}  

\begin{document}
\pagestyle{headings}
\mainmatter
\def\ECCVSubNumber{2310}  

\title{Ultra-high-resolution unpaired stain transformation via Kernelized Instance Normalization} 


\titlerunning{Ultra-high-resolution unpaired stain transformation}
%
\author{Ming-Yang Ho\thanks{%
    Corresponding author} \and
Min-Sheng Wu \and
Che-Ming Wu}
\authorrunning{Ho. et al.}
%
\institute{aetherAI, Taipei, Taiwan\\
\email{\{kaminyouho,vincentwu,uno\}@aetherai.com}\\
\url{https://www.aetherai.com/}}
\maketitle

\begin{abstract}
While hematoxylin and eosin (H\&E) is a standard staining procedure, immunohistochemistry (IHC) staining further serves as a diagnostic and prognostic method. However, acquiring special staining results requires substantial costs.
Hence, we proposed a strategy for ultra-high-resolution unpaired image-to-image translation: Kernelized Instance Normalization (KIN), which preserves local information and successfully achieves seamless stain transformation with constant GPU memory usage. Given a patch, corresponding position, and a kernel, KIN computes local statistics using convolution operation. In addition, KIN can be easily plugged into most currently developed frameworks without re-training.
We demonstrate that KIN achieves state-of-the-art stain transformation by replacing instance normalization (IN) layers with KIN layers in three popular frameworks and testing on two histopathological datasets. Furthermore, we manifest the generalizability of KIN with high-resolution natural images. Finally, human evaluation and several objective metrics are used to compare the performance of different approaches.
Overall, this is the first successful study for the ultra-high-resolution unpaired image-to-image translation with constant space complexity. Code is available at: \url{https://github.com/Kaminyou/URUST}.

\end{abstract}

\section{Introduction}
\begin{figure}[!htb]
\centering
\resizebox{0.88\textwidth}{!}{%
\includegraphics[]{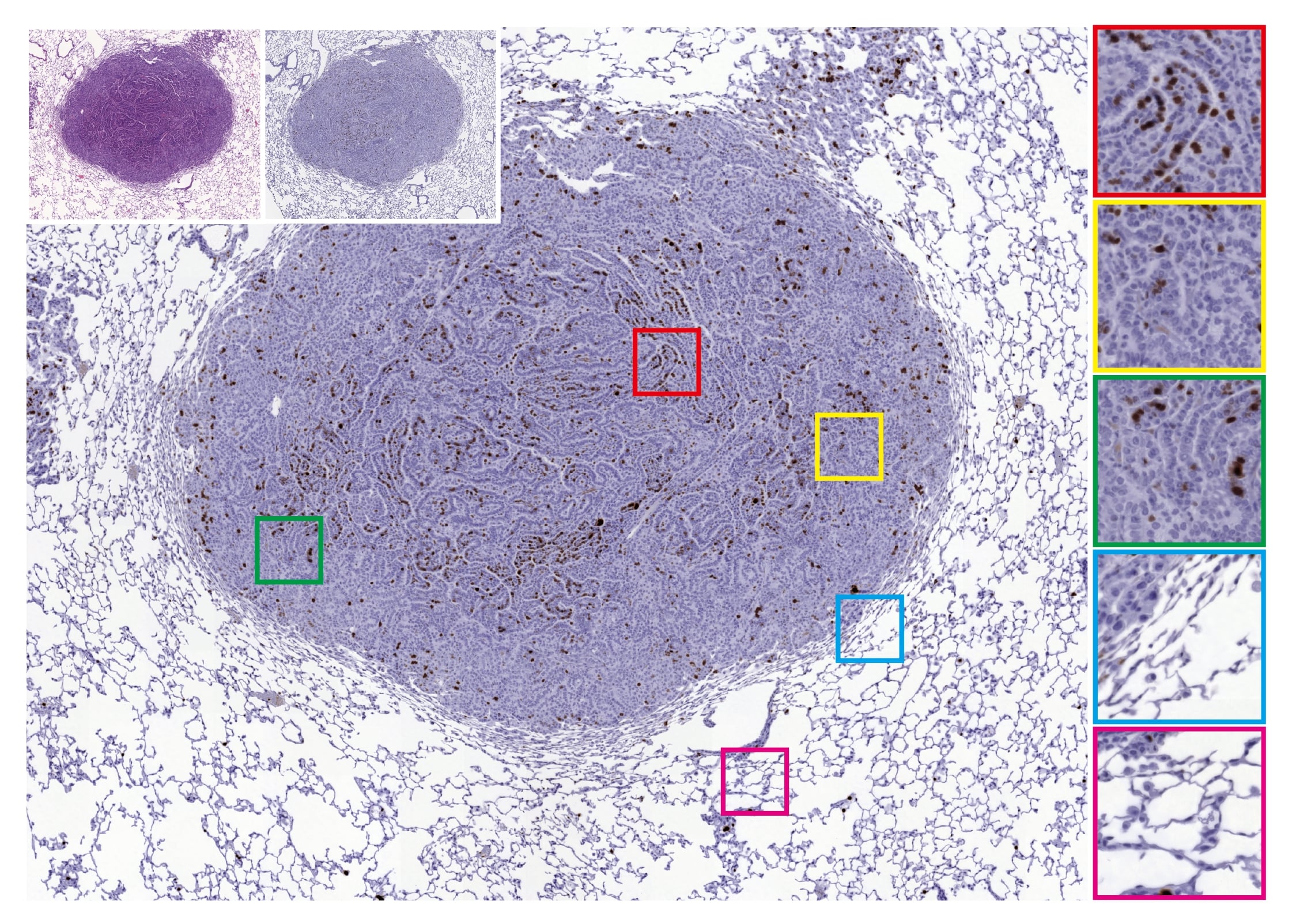}
}
\caption{\textbf{An ultra-high-resolution translated result ($\mathbf{7,328 \times 8,899}$ pixels) from our Kernelized Instance Normalization (KIN).} The whole slide image (WSI) was translated from source stain to target stain (on the upper left) with constant space complexity (GPU memory) via KIN, and local appearance was preserved. On the right side, five close-ups demonstrate the detail.}
\label{fig:best}
\end{figure}

\afterpage{%
\begin{figure}[t!]
\centering
\resizebox{0.66\textwidth}{!}{%
\includegraphics[]{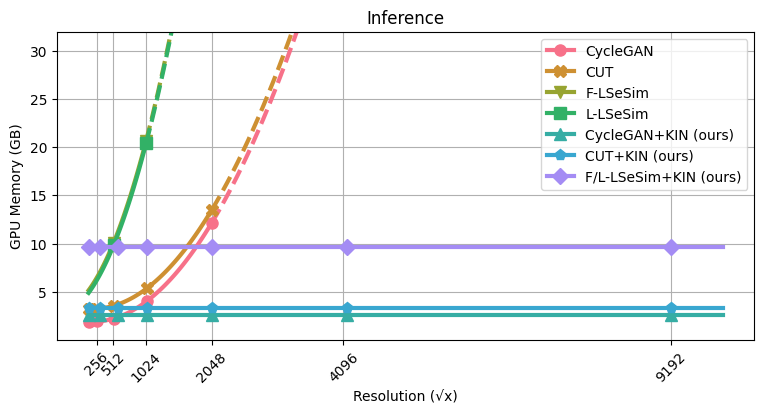}
}
\caption{\textbf{Comparison of GPU memory usage among different unpaired image-to-image translation approaches.}
Compared with the models using Instance Normalization (IN), which has limitation (marked by the dashed line) on a 32G GPU (NVIDIA V-100), our Kernelized Instance Normalization (KIN) approach can translate an ultra-high-resolution image with constant GPU memory usage (less than 5GB).}
\label{fig:inference_usage}
\end{figure}
}
Histological staining, highlighting cellular components with dyes, is crucial in clinical diagnosis~\cite{alturkistani2016histological}, which enables visualization of cells and extracellular matrix and abnormal identification. Since specific cellular components or biomarkers can be distinguished when particular dyes attach specific molecules in tissues, different staining methods are applied to diagnose various diseases and their subtypes~\cite{inamura2018update,fragomeni2018molecular,birkman2018gastric}.
The standard stain (or routine stain) is hematoxylin and eosin (H\&E). While hematoxylin stains nuclei, eosin can stain cytoplasm. Immunohistochemistry (IHC) protocol is further developed to detect the presence of specific protein markers. For example, Ki-67 and ER staining can quantify the presence of Ki-67 and ER biomarkers, respectively. In clinical practice, high Ki-67 expression is considered a poor prognostic factor~\cite{luo2019increased}, while the presence of ER indicates the suitability of choosing specific target therapies that benefit related disease subtypes~\cite{oshi2020degree}.
However, compared with H\&E staining, the IHC staining process is much more expensive and requires extra biopsies, which are limited materials. With the development of deep learning-based image-to-image translation, virtually translating H\&E into different IHC staining can be achieved. For example, de Haan \etal. proposed a supervised deep learning approach via CycleGAN~\cite{CycleGAN2017} to transform stained images from H\&E to Masson’s Trichrome (MT) staining~\cite{de2021deep}. While supervised pair-wise training is desirable, this approach requires perfectly paired staining images, necessitating de-staining and re-staining processes, and is not practically efficient in a clinical scenario. 
Most datasets are composed of unpaired H\&E and IHC images from consecutive sections. Several methodologies have been proposed and successfully tackled the unpaired image-to-image translation problem~\cite{pang2021image}. Regardless of the astonishing performance, the existing methods are limited to low-resolution images and rarely explore the images with ultra-high-resolution. In histopathology, whole slide images (WSIs) are usually larger than $10,000\times 10,000$ pixels. The main challenge of transforming a WSI is the limitation of GPU memory capacity.
Patch-wise inference with an assembly process can tackle ultra-high-resolution image-to-image translation, but tiling artifacts between adjacent patches would be a critical problem. Traditionally, overlapping windows have been leveraged to smooth the transitions but have limited effectiveness. Considering the mean and standard deviation calculated in instance normalization (IN) layers might influence hue and contrast, recently, Chen \etal developed Thumbnail Instance Normalization (TIN) for both ultra-high-resolution style transfer as well as image-to-image translation tasks~\cite{chen2022towards}. Unfortunately, while their approach could overcome the resolution limitation, their erroneous assumption that all patches share global mean and standard deviation would lead to dramatically over/under-colorizing according to our comprehensive experiment, which is confirmed in Section~\ref{sec:results}.
To compensate for all the above limitations, we proposed a Kernelized Instance Normalization (KIN) layer that can replace the original IN layer during the inference process without re-training the models. With the help of KIN, images with arbitrary resolution can be translated with constant GPU memory space (as demonstrated in Fig.~\ref{fig:best} and ~\ref{fig:inference_usage}). Moreover, utilizing the statistics of neighboring patches instead of global ones like TIN, our approach can further preserve the hue and contrast locally, which is especially paramount in stain transformation tasks. Besides the translation of H\&E to four IHC staining, we additionally demonstrated the generalizability of KIN with natural images by translating summer to autumn style.
Our novel contribution can be summarized as follows:
\begin{itemize}
     \item To the best of our knowledge, this is the first successful study for the ultra-high-resolution unpaired image-to-image translation with constant space complexity (GPU memory), which manifests state-of-the-art outcomes in stain transformation and can also be generalized to natural images.
    \item Without re-training the models, our KIN module can be seamlessly inserted into most currently developed frameworks that have IN layers, such as CycleGAN~\cite{CycleGAN2017}, CUT~\cite{park2020contrastive}, and LSeSim~\cite{zheng2021spatiallycorrelative}.
    \item With the KIN module, local contrast and hue information in translated images can be well preserved. Besides, different kernels can be further applied to subtly adjust the translated images.
\end{itemize}

\section{Related works}

\subsection{Unpaired image-to-image translation}
Several frameworks have been proposed for unpaired image-to-image translation. CycleGAN~\cite{CycleGAN2017}, DiscoGAN~\cite{kim2017learning}, and DualGAN~\cite{yi2017dualgan} were first presented to overcome the supervised pairing constraint via cycle consistency. However, subtle information was forced to be retained in the translated image to achieve better reconstruction, causing detrimental effects when two domains are substantially different such as dog-to-cat translation. Besides, a reverse mapping function might not always exist, which inevitably leads to artifacts in translated images. Recently, strategies beyond cyclic loss have been developed to reach one-sided unpaired image-to-image translation. While DistanceGAN~\cite{Benaim2017OneSidedUD} enforced distance consistency between different parts of the same sample in each domain, CUT~\cite{park2020contrastive} leveraged contrastive loss to maximize the patch-wise similarity between domains and achieved remarkable results. LSeSim~\cite{zheng2021spatiallycorrelative} further utilized spatially correlation to maximize structural similarity and eliminate the domain-specific features.

\subsection{Image-to-image translation for stain transformation}
Transforming one stained tissue into another specific stain will dramatically save laboratory resources and money. Hence, growing research has leveraged unsupervised image-to-image translation to conduct stain transformation in several medical scenarios. Levy \etal. translated H\&E to trichrome staining via CycleGAN for liver fibrosis staging~\cite{levy2020preliminary}. Kapil \etal translated Cytokeratin to PD-L1 staining to bypass re-staining for segmentation~\cite{kapil2019dasgan}. de Haan \etal translated H\&E to Masson’s Trichrome, periodic acid-Schiff, and Jones silver stain for improving preliminary diagnosis for kidney diseases via CycleGAN with perfectly paired images~\cite{de2021deep}. Lahiani \etal further broke the limitation of $256 \times 256$-pixel image patches by applying perceptual embedding consistency for H\&E to FAP-CK transformation~\cite{lahiani2020seamless}. However, the lack of detailed description hampers the implementation of their methodology.

\subsection{Ultra-high resolution image-to-image translation}
Ultra-high-resolution images are ubiquitous in photography, artwork, posters, ultra-high (\eg 8K) videos, and especially, WSIs in digital pathology (usually, larger than $10,000 \times 10,000$). Due to the massive computational costs, transforming these images into different styles will be difficult. Traditionally, strategies have been proposed to address the problem of tiling artifacts created by patch-wise-based methods, including utilizing a larger overlapping window~\cite{lahiani2019virtualization} or freezing IN layers during testing time~\cite{de2018stain}. By providing a patch-wise style transfer network with Thumbnail Instance Normalization (TIN), Chen \etal performed ultra-high-resolution style transformation with constant GPU memory usage~\cite{chen2022towards}. Also, they applied their framework to an image-to-image translation task. However, according to our experiment, TIN may result in over/under-colorizing for their fallacious assumption that all patches can be normalized with the same global mean and standard deviation. 

\begin{figure}[t!]
\centering
\resizebox{\textwidth}{!}{%
\includegraphics[]{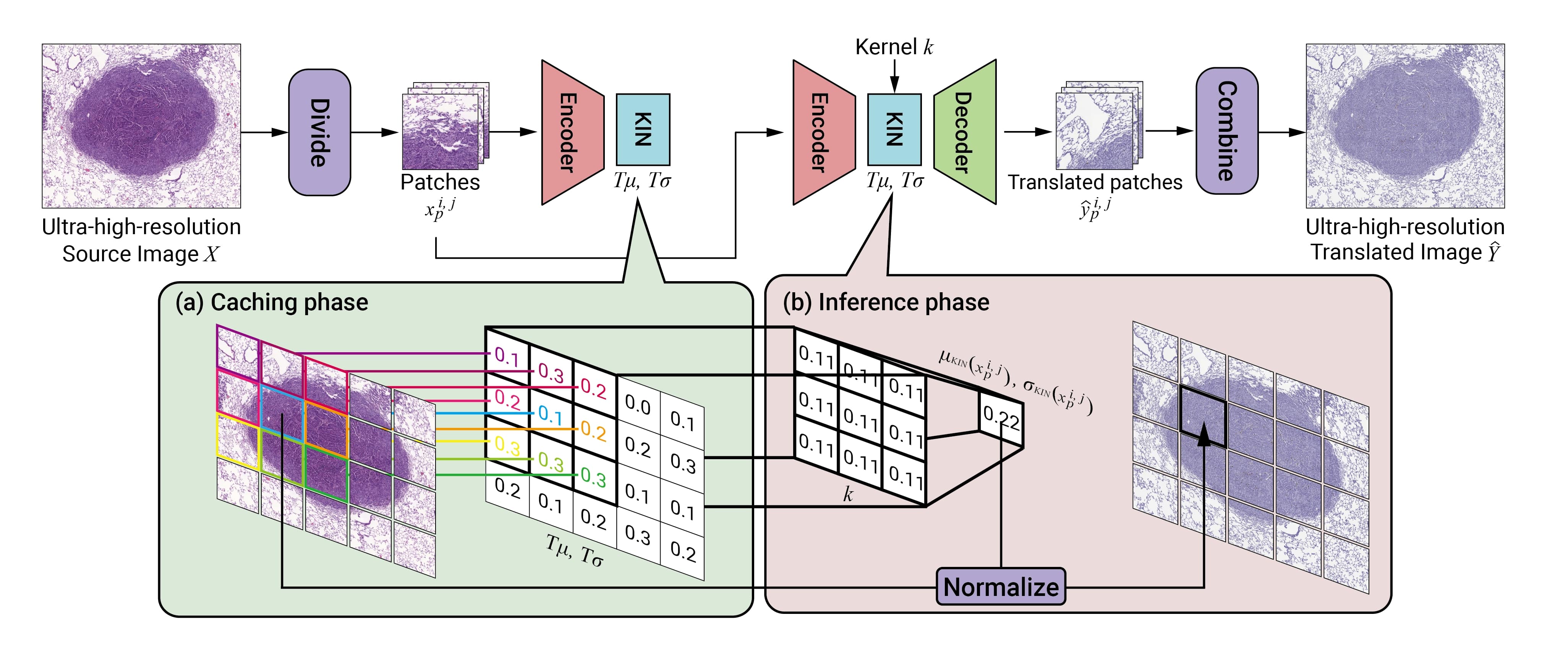}
}
\caption{\textbf{Overall framework of proposed method.} An ultra-high-resolution H\&E image is passed through the caching phase and inference phase to be translated into an ultra-high-resolution IHC image. (a) In the caching phase, all mean $\mu$ and standard deviation $\sigma$ values of patches will be cached in caching tables $T_{\mu}$ and $T_{\sigma}$ by the Kernelized Instance Normalization (KIN) layer. (b) In the inference phase, a kernel $k$ will convolute the caching table to compute $\mu_{KIN}$ and $\sigma_{KIN}$ for instance normalization. Taking the neighboring statistics into account, our method can preserve local appearance.
}
\label{fig:method}
\end{figure}

\section{Proposed method}

\subsection{Overall Framework}
Our framework targets one-sidedly translating an ultra-high-resolution image $X$ in domain $\mathcal{X}$ (e.g., H\&E stain domain) into image $\hat{Y}$ in domain $\mathcal{Y}$ (e.g., IHC domain), in which $X, \hat{Y} \in R^{H \times W \times C}$, $H$ and $W$ are the height and width of $X$, via a mapping function, generator $\mathcal{G}$. 

\begin{equation}
    \hat{Y} = \mathcal{G}(X), \mathcal{G}:\mathcal{X} \rightarrow \mathcal{Y}
\end{equation}

Collections of unpaired $\textbf{X}$ in $\mathcal{X}$ and $\textbf{Y}$ in $\mathcal{Y}$ would be first cropped into patches with the size of $512 \times 512$ pixels to train a generator $\mathcal{G}$. 
As our KIN module is only applied during the testing time and can be inserted into any framework with IN layers, we followed the original training process and hyperparameters proposed in the paper of CycleGAN~\cite{CycleGAN2017}, CUT~\cite{park2020contrastive}, and LSeSim~\cite{zheng2021spatiallycorrelative} to train the corresponding generators with their specific designed losses without any modification.
During the testing process, all the IN layers in $\mathcal{G}$ are replaced  with KIN layers. Given an image $X$, non-overlapped patches $x^{i, j}_{p}$ are cropped with the size of $512\times 512$. The coordinates $i, j$ of each patch $x^{i, j}_{p}$ corresponding to the original $X$ would be recorded simultaneously. For example, an $M \times N$ image would be cropped into $\lfloor M / 512 \rfloor \times \lfloor N / 512 \rfloor$ patches with coordinates of \{$0, 1, ..., \lfloor M / 512 \rfloor $\} $\times$ \{$0, 1, ..., \lfloor N / 512 \rfloor $\}. Two caching tables of size $\lfloor M / 512 \rfloor \times \lfloor N / 512 \rfloor \times C$, in which $C$ denotes the number of channels, would be initialized in each KIN for caching mean and standard deviation calculated. 
As illustrated in Fig~\ref{fig:method}, we divide the testing process into two phases: caching and inference. During caching phase, each patch ${x^{i, j}_{p}}$ with its corresponding coordinates $i$, $j$ are the input of the generator $\mathcal{G}$, and the calculated mean $\mu(x^{i, j}_{p})$ and standard deviation $\sigma(x^{i, j}_{p})$ after passing the KIN will be cached.
During the inference phase, ${x^{i, j}_{p}}$, its corresponding coordinates $i$, $j$ and a kernel $k$ are the input of the generator $\mathcal{G}$. The kernel $k \in R^{h \times w}$, where $h$ and $w$ are the height and width of $k$, is adjustable.
When passing through the KIN layer, a region with the same size of kernel $k$ extended from ${i, j}$ will be extracted from the caching table and convolute with kernel k to compute mean $\mu_{KIN}(x^{i, j}_{p})$ and standard deviation $\sigma_{KIN}(x^{i, j}_{p})$ which are used to normalize the feature maps. 
All the cropped patches will be passed to the $\mathcal{G}$ in the aforementioned manner to yield translated patches $\hat{y}^{i, j}_{p}$. Eventually, all $\hat{y}^{i, j}_{p}$ are assembled into an ultra-high-resolution translated image $\hat{Y}$.

\subsection{Kernelized Instance Normalization (KIN)}

IN~\cite{ulyanov2016instance} has been widely used in GAN-based models for image generation and dramatically improved image quality~\cite{pang2021image}. Besides, multiple styles can be obtained by conditionally replacing the $\mu$ and $\sigma$ in the IN layer~\cite{dumoulin2017learned}. IN can be formulated by:

\begin{equation}
    IN(X) = \gamma(\frac{X-\mu(X)}{\sigma(X)}) + \beta
\end{equation}

For each instance in a batch, $\mu(X)$ and $\sigma(X)$ are calculated in a channel-wise manner, in which $\mu(X), \sigma(X) \in R^{B \times C}$, $B$ denotes the batch size and $\gamma$ and $\beta$ are trainable parameters. 
We hypothesize that adjacent patches share similar statistics including the $\mu$ and $\sigma$ computed in IN, and thus proposed KIN that could further alleviate the subtle incongruity that induces the tiling artifacts when adjacent patches are assembled.
KIN is the extension of the original IN layer with extra two caching tables $T_{\mu}$ and $T_{\sigma}$ to spatially store $\mu(X)$ and $\sigma(X)$ values and additionally supports convolution operation on the caching tables with a given kernel $k$. During the caching phase, KIN input a cropped patch $x^{i, j}_p$ with its spatial information, $i, j$. $\mu(x^{i, j}_p)$ and $\sigma(x^{i, j}_p)$ are computed as the original IN and cached.

\begin{equation}
    T_{\mu}[i,j] := \mu(x^{i, j}_p), x^{i, j}_p \text{ is cropped from } X_{i,j}
\end{equation}
\begin{equation}
    T_{\sigma}[i,j] := \sigma(x^{i, j}_p), x^{i, j}_p \text{ is cropped from } X_{i,j}
\end{equation}

During the inference phase, given a kernel $k$ with the size of $2q+1$, $\mu_{KIN}(x^{i,j}_p)$ and $\sigma_{KIN}(x^{i,j}_p)$ are computed by convoluting $k$ on cache tables to generate translated images. To address the boundary cases, the cache tables would be padded initially with edge values.

\begin{equation}
    \mu_{KIN}(x^{i, j}_{p}) = \sum_{u=-q}^{q}\sum_{v=-q}^{q}T_{\mu}[i+u,j+v]\cdot	K[q+u, q+v], \forall i,j
\end{equation}
\begin{equation}
    \sigma_{KIN}(x^{i, j}_{p}) = \sum_{u=-q}^{q}\sum_{v=-q}^{q}T_{\sigma}[i+u,j+v]\cdot	K[q+u, q+v], \forall i,j
\end{equation}
\begin{equation}
    KIN(x^{i,j}_p, i, j) = \gamma(\frac{x^{i,j}_p - \mu_{KIN}(x^{i, j}_{p})}{\sigma_{KIN}(x^{i, j}_{p})}) + \beta
\end{equation}

\section{Datasets}

\subsection{Automatic Non-rigid Histological Image Registration (ANHIR)}

Automatic Non-rigid Histological Image Registration (ANHIR) dataset\linebreak ~\cite{borovec2020anhir,borovec2018benchmarking,bueno2019aidpath,fernandez2002system,gupta2018stain,mikhailov2018immune} consists of high-resolution WSIs from different tissue samples (lesions, lung lobes, breast tissue, kidney tissue, gastric tissue, colon adenocarcinoma, and mammary gland). The acquired images are organized in sets of consecutive tissue slices stained by various dyes, including H\&E, Ki-67, ER/PR, CD4/CD8/CD68, etc., with sizes vary from $15,000 \times 15,000$ to $50,000 \times 50,000$ pixels.
We randomly sampled three types of tissues to conduct our experiments. Each experiment comprises H\&E stain and one target IHC stain: breast tissue (from H\&E to PR), colon adenocarcinoma (COAD) (from H\&E to CD4\&CD68), and lung lesion (from H\&E to Ki-67).

\subsection{Glioma}

The private glioma dataset was collected from H\&E ($98,304 \times 93,184$ pixels) and epidermal growth factor receptor (EGFR) IHC ($102,400 \times 93,184$ pixels) stained tissue microarrays, and each comprised 105 tissue samples corresponding to 105 different patients. 
Totally 105 H\&E stained tissue images with their consecutive EGFR counterparts were cropped from the microarrays and the image sizes vary from $7,000 \times 7,000$ to $10,000 \times 10,000$ pixels. We randomly selected 55 samples as the training set while the other 50 pairs as the testing set.

\subsection{Kyoto summer2autumn}

An extra natural image dataset was used to validate the generalizability of our methodology. We collected 17 and 20 high-resolution ($3456 \times 5184$ pixels) unpaired images taken in Tokyo during summer and autumn, respectively, as the training set and additional four summer images were used as a testing set. This Kyoto summer2autumn dataset\footnote{Kyoto summer2autumn dataset is available at: \url{https://github.com/Kaminyou/Kyoto-summer2autumn}} was released to facilitate solving ultra-high-resolution-related problems that most computer vision studies might encounter.

\section{Experiments}

\subsection{Experimental settings}
Three popular unpaired image-to-image translation frameworks: CycleGAN~\cite{CycleGAN2017}, CUT~\cite{park2020contrastive}, and L-LSeSim~\cite{zheng2021spatiallycorrelative}, were utilized to verify our approach. We followed the hyperparameter settings described in the original papers during the training process except for the model output size, which was changed from $256 \times 256$ to $512 \times 512$. We trained the CycleGAN, CUT, and L-LSeSim for 50, 100, and 100 epochs.
Models were trained and tested on three datasets: ANHIR, glioma, and Kyoto summer2autumn. Due to the insufficiency of WHIs in ANHIR dataset, we could only inference on the training set (note that training was in an unsupervised manner) while glioma and Kyoto summer2autumn datasets can be further split into training and testing sets.
We replaced all IN layers with KIN layers in the generators during the inference process. One ultra-high-resolution image would be cropped into non-overlapped patches and pass through the KIN module. Translated patches were assembled to the final translated output. Constant and Gaussian kernels with sizes of 3, 7, and 11 were used to generate the best results.
Translated images generated with KIN were compared with those from IN and TIN. Due to the GPU memory limitation, translated images generated with IN were also in a patch-wise  (512$\times$512) manner, which is the same as the patch-wise IN in Chen \etal's work~\cite{chen2022towards}.

\subsection{Metrics}
In addition to the visualization of the translated images, we calculated Fréchet inception distance (FID)~\cite{heusel2017gans}, histogram correlation, Sobel gradients~\cite{irwin1968isotropic} in YCbCr color domain, perception image quality evaluator (PIQE)~\cite{venkatanath2015blind}, and natural image quality evaluator (NIQE)~\cite{mittal2012making} to comprehensively evaluate the quality of translated ultra-high-resolution images.
However, due to the limitations of the available metrics that the tiling artifacts are difficult to be fairly graded and the unavailability of the perfectly matched counterpart, we conducted two human evaluation studies with five specialists:
(a) quality challenge: given one source, one reference, and three translated images generated by patch-wise IN, TIN, and our KIN methods, respectively, specialists were asked to select the best among three translated images in 30 seconds.
Since the images generated by CycleGAN and L-LSeSim were atrocious, we only chose images generated by CUT;
(b) fidelity challenge: given one real image and one translated image, specialists were asked to select the one which is personally considered realistic in 10 seconds. We followed the protocol of the AMT perceptual studies from
Isola \etal~\cite{isola2017image} but adjusted the time limitation as our images are extremely large.
Since the data in ANHIR breast, COAD, and lung lesion subdatasets are insufficient, we combined these subdatasets as single ANHIR dataset and randomly selected pairs of real and translated WSIs from it.

\subsection{Results} \label{sec:results}

\begin{figure}[!htb]
\centering
\resizebox{0.845\textwidth}{!}{%
\includegraphics[]{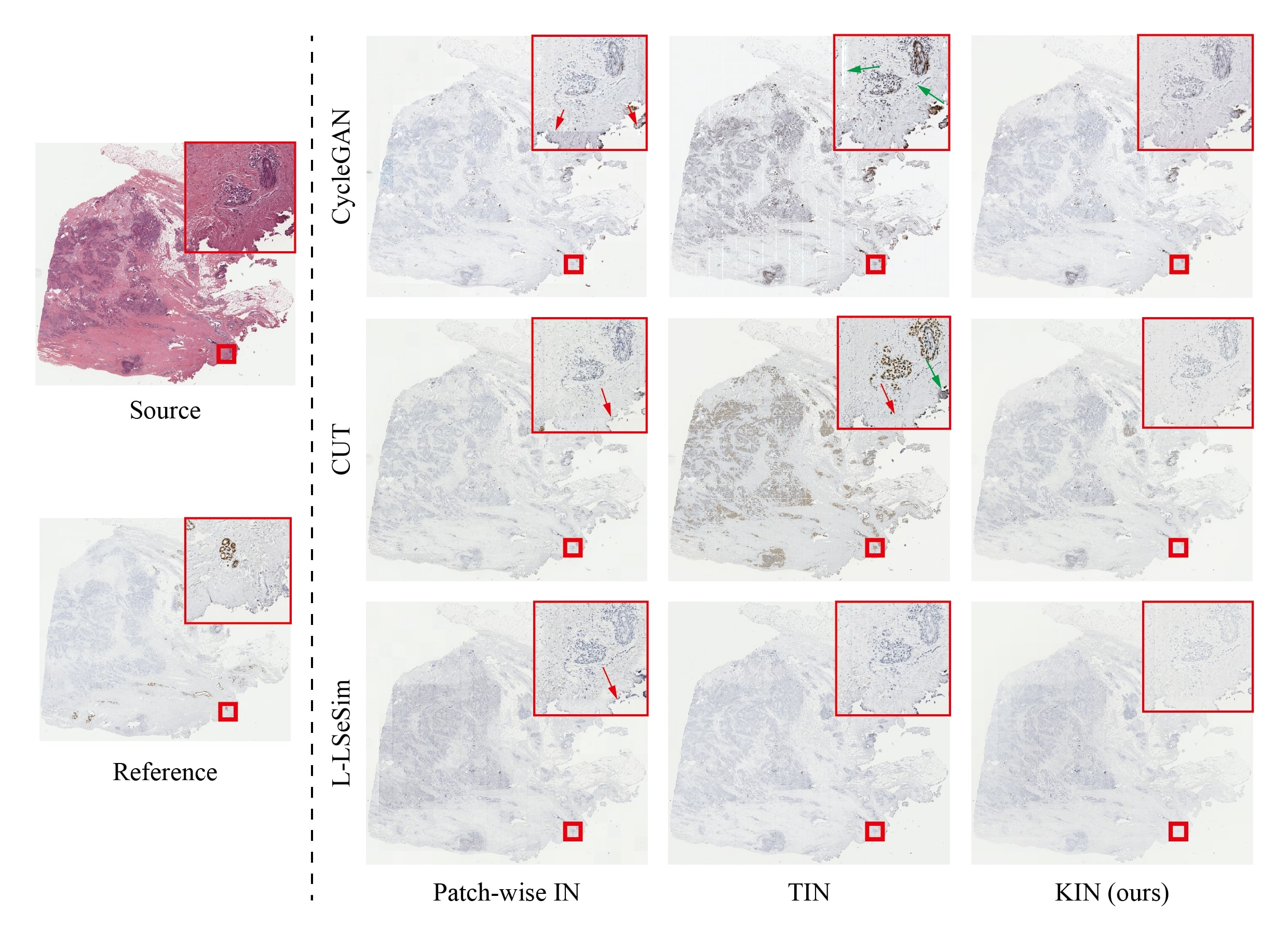}
}
\caption{\textbf{H\&E-to-PR stain transformation results on ANHIR breast dataset ($\mathbf{10,205 \times 10,933}$ pixels)} generated by different frameworks with IN, TIN, and KIN layers. Red arrows indicate tiling artifacts; green arrows indicate over/under-colorizing. CUT+KIN achieved the best performance. Zoom in for better view.} 
\label{fig:breast}
\end{figure}

\begin{figure}[!htb]
\centering
\resizebox{0.845\textwidth}{!}{%
\includegraphics[]{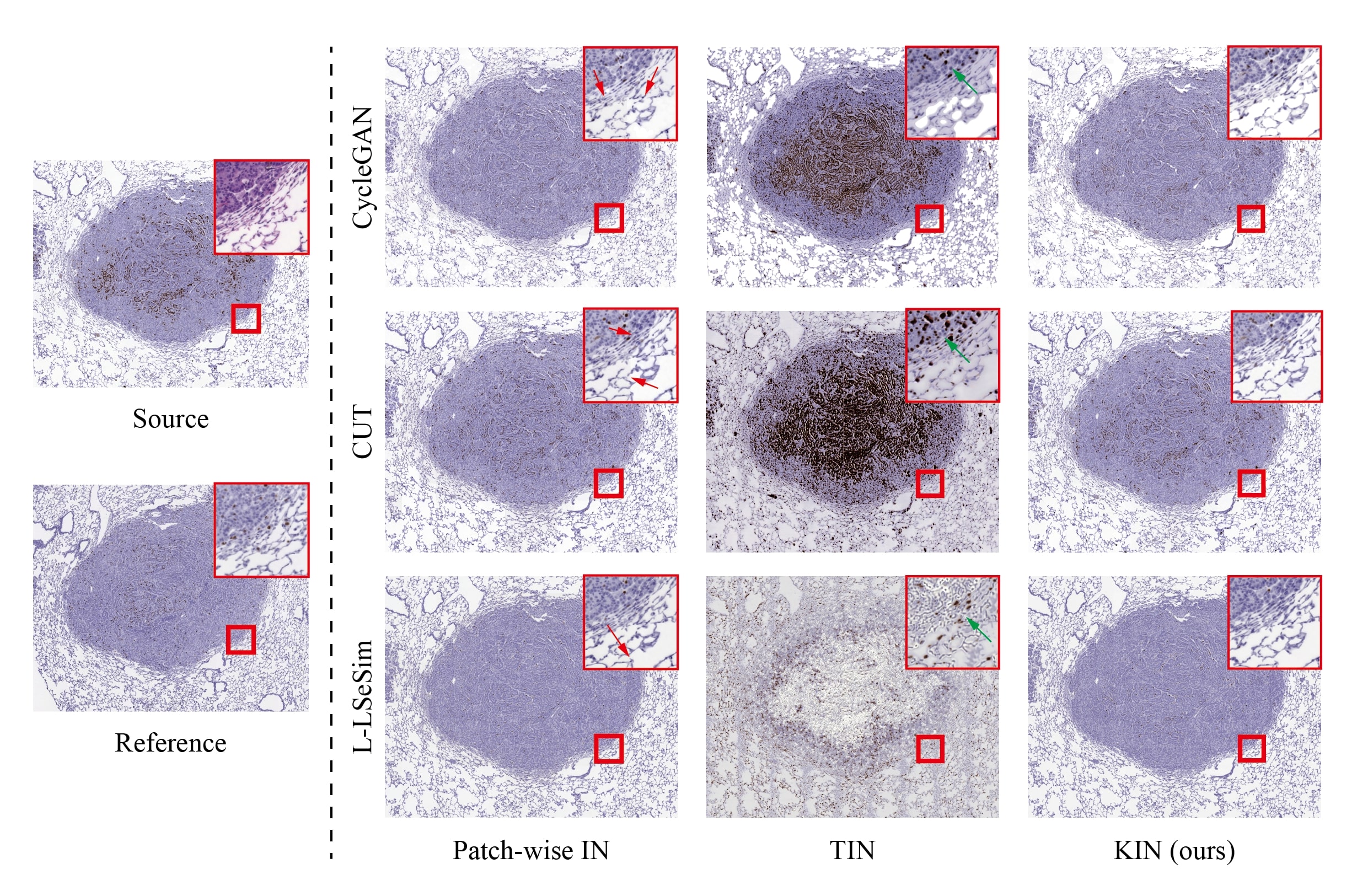}
}
\caption{\textbf{H\&E-to-Ki-67 stain transformation results on ANHIR lung lesion dataset ($\mathbf{7,336 \times 8,915}$ pixels)} generated by different frameworks with IN, TIN, and KIN layers. Red arrows indicate tiling artifacts; green arrows indicate over/under-colorizing. CUT+KIN achieved the best performance. Zoom in for better view.}
\label{fig:lung_lesion}
\end{figure}

\begin{figure}[!htb]
\centering
\resizebox{\textwidth}{!}{%
\includegraphics[]{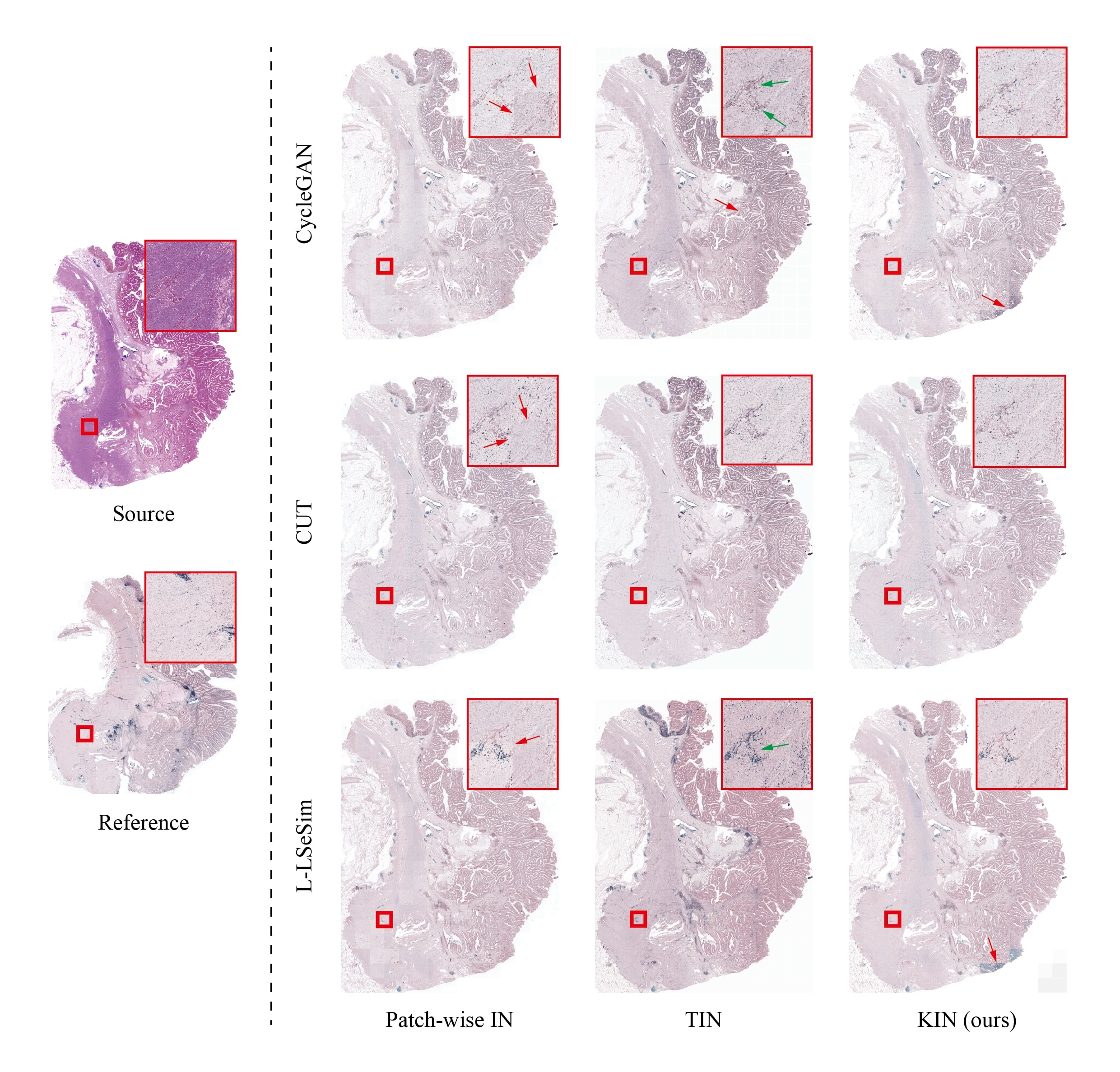}
}
\caption{\textbf{H\&E-to-CD4\&CD8 stain transformation results on ANHIR COAD dataset ($\mathbf{9,816 \times 8,433}$ pixels)} generated by different frameworks with IN, TIN, and KIN layers. Red arrows indicate tiling artifacts; green arrows indicate over/under-colorizing. CUT+KIN achieved the best performance. Zoom in for better view.}
\label{fig:COAD}
\end{figure}

\begin{figure}[!htb]
\centering
\resizebox{0.91\textwidth}{!}{%
\includegraphics[]{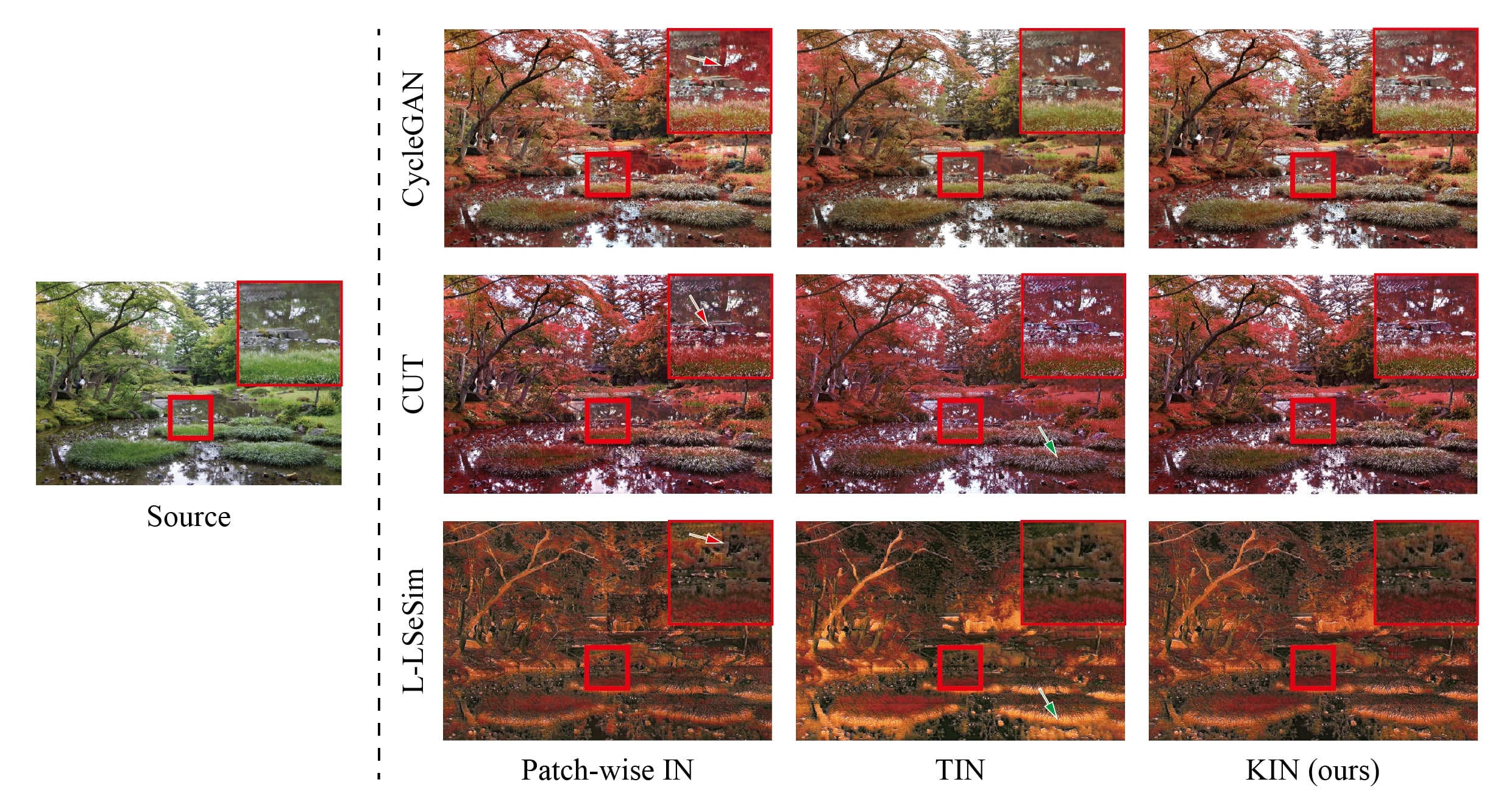}
}
\caption{\textbf{Image-to-image translation results on Kyoto summer2autumn testing set ($\mathbf{3,456 \times 5,184}$ pixels)} generated by different frameworks with IN, TIN, and KIN layers. Red arrows indicate tiling artifacts; green arrows indicate over/under-colorizing. CUT+KIN achieved the best performance. Zoom in for better view.}
\label{fig:kyoto_test}
\end{figure}

\begin{table}[!htb]
\centering
\caption{\textbf{Quantitative results for ANHIR dataset.}  For each experiment, the bold shows the best performance; the underline indicates that KIN surpasses IN and TIN.}
\resizebox{\textwidth}{!}{%
\begin{threeparttable}
\label{tab:ANHIR}
\begin{tabular}{llrrrrrrrrrrrrrrr}
\toprule
 &
  \textbf{} &
  \multicolumn{5}{c}{\textbf{Breast}} &
  \multicolumn{5}{c}{\textbf{COAD}} &
  \multicolumn{5}{c}{\textbf{Lung lesion}} \\
  \cmidrule(l){3-7} \cmidrule(l){8-12} \cmidrule(l){13-17}
 &
  \textbf{} &
  \multicolumn{1}{c}{\textbf{FID}$\downarrow$} &
  \multicolumn{1}{c}{\textbf{Corr.}$\uparrow$} &
  \multicolumn{1}{c}{\textbf{Grad.}$\downarrow$} &
  \multicolumn{1}{c}{\textbf{PIQE}$\downarrow$} &
  \multicolumn{1}{c}{\textbf{NIQE}$\downarrow$} &
  \multicolumn{1}{c}{\textbf{FID}$\downarrow$} &
  \multicolumn{1}{c}{\textbf{Corr.}$\uparrow$} &
  \multicolumn{1}{c}{\textbf{Grad.}$\downarrow$} &
  \multicolumn{1}{c}{\textbf{PIQE}$\downarrow$} &
  \multicolumn{1}{c}{\textbf{NIQE}$\downarrow$} &
  \multicolumn{1}{c}{\textbf{FID}$\downarrow$} &
  \multicolumn{1}{c}{\textbf{Corr.}$\uparrow$} &
  \multicolumn{1}{c}{\textbf{Grad.}$\downarrow$} &
  \multicolumn{1}{c}{\textbf{PIQE}$\downarrow$} &
  \multicolumn{1}{c}{\textbf{NIQE}$\downarrow$} \\ \midrule
\multicolumn{1}{l|}{\multirow{3}{*}{CycleGAN}} &
  IN\mbox{*} &
  98.60 &
  -0.07 &
  13.62 &
  \textbf{4.95} &
  9.39 &
  103.25 &
  75.25 &
  15.08 &
  5.26 &
  \textbf{9.08} &
  \textbf{76.15} &
  -4.49 &
  9.92 &
  \textbf{62.69} &
  13.14 \\
\multicolumn{1}{l|}{} &
  TIN &
  179.14 &
  -28.91 &
  14.37 &
  6.16 &
  9.56 &
  \textbf{100.78} &
  \textbf{79.53} &
  16.68 &
  15.79 &
  9.55 &
  239.19 &
  \textbf{17.46} &
  \textbf{9.53} &
  67.79 &
  \textbf{11.96} \\
\multicolumn{1}{l|}{} &
  KIN &
  \textbf{\underline{96.09}} &
  \textbf{\underline{11.53}} &
  \textbf{\underline{12.93}} &
  5.29 &
  \textbf{\underline{7.36}} &
  108.32 &
  43.60 &
  \textbf{\underline{14.95}} &
  \textbf{\underline{5.15}} &
  9.69 &
  103.48 &
  -2.16 &
  9.94 &
  63.81 &
  12.16 \\ \midrule
\multicolumn{1}{l|}{\multirow{3}{*}{CUT}} &
  IN\mbox{*} &
  \textbf{71.00} &
  \textbf{35.56} &
  14.55 &
  \textbf{3.00} &
  12.15 &
  95.87 &
  74.64 &
  14.76 &
  4.50 &
  8.96 &
  \textbf{54.86} &
  -5.79 &
  \textbf{10.41} &
  58.32 &
  12.20 \\
\multicolumn{1}{l|}{} &
  TIN &
  125.18 &
  39.50 &
  17.04 &
  3.42 &
  10.98 &
  \textbf{91.81} &
  33.29 &
  15.49 &
  11.96 &
  9.02 &
  251.48 &
  \textbf{80.14} &
  11.80 &
  \textbf{32.08} &
  12.20 \\
\multicolumn{1}{l|}{} &
  KIN &
  72.59 &
  36.32 &
  \textbf{\underline{14.05}} &
  3.27 &
  \textbf{\underline{10.66}} &
  93.68 &
  \textbf{\underline{76.45}} &
  \textbf{\underline{14.60}} &
  \textbf{\underline{4.49}} &
  \textbf{\underline{8.94}} &
  56.38 &
  -9.63 &
  10.58 &
  60.13 &
  \textbf{\underline{12.08}} \\ \midrule
\multicolumn{1}{l|}{\multirow{3}{*}{L-LSeSim}} &
  IN\mbox{*} &
  \textbf{65.82} &
  31.57 &
  15.04 &
  \textbf{3.03} &
  13.36 &
  100.42 &
  48.15 &
  13.64 &
  \textbf{4.23} &
  8.45 &
  \textbf{56.30} &
  -3.82 &
  \textbf{9.71} &
  46.40 &
  13.88 \\
\multicolumn{1}{l|}{} &
  TIN &
  89.94 &
  22.16 &
  \textbf{12.75} &
  3.29 &
  13.18 &
  100.50 &
  41.37 &
  15.67 &
  9.56 &
  \textbf{8.14} &
  231.13 &
  \textbf{59.71} &
  12.68 &
  \textbf{44.29} &
  \textbf{11.13} \\
\multicolumn{1}{l|}{} &
  KIN &
  67.46 &
  \textbf{\underline{31.58}} &
  14.31 &
  3.19 &
  \textbf{\underline{12.35}} &
  \textbf{\underline{100.04}} &
  \textbf{\underline{51.62}} &
  \textbf{\underline{13.34}} &
  4.44 &
  8.22 &
  62.74 &
  -4.77 &
  9.91 &
  47.99 &
  13.48 \\ \bottomrule
\end{tabular}%
\begin{tablenotes}
  \footnotesize
  \item IN\mbox{*}: Patch-wise IN; Corr.: Histogram correlation; Grad.: Sobel gradients; $\downarrow$: the lower the better; $\uparrow$: the higher the better.
\end{tablenotes}
\end{threeparttable}
}
\end{table}

\begin{table}[!htb]
\centering
\caption{\textbf{Quantitative results for Glioma dataset.}  For each experiment, the bold shows the best performance; the underline indicates that KIN surpasses IN and TIN.}
\resizebox{0.95\textwidth}{!}{%
\begin{threeparttable}
\label{tab:glioma}
\begin{tabular}{llrrrrrrrrrr}
\toprule
 &
  \textbf{} &
  \multicolumn{5}{c}{\textbf{Glioma (training set)}} &
  \multicolumn{5}{c}{\textbf{Glioma (testing set)}} \\ 
  \cmidrule(l){3-7} \cmidrule(l){8-12} 
 &
  \textbf{} &
  \multicolumn{1}{c}{\textbf{FID}$\downarrow$} &
  \multicolumn{1}{c}{\textbf{Corr.}$\uparrow$} &
  \multicolumn{1}{c}{\textbf{Grad.}$\downarrow$} &
  \multicolumn{1}{c}{\textbf{PIQE}$\downarrow$} &
  \multicolumn{1}{c}{\textbf{NIQE}$\downarrow$} &
  \multicolumn{1}{c}{\textbf{FID}$\downarrow$} &
  \multicolumn{1}{c}{\textbf{Corr.}$\uparrow$} &
  \multicolumn{1}{c}{\textbf{Grad.}$\downarrow$} &
  \multicolumn{1}{c}{\textbf{PIQE}$\downarrow$} &
  \multicolumn{1}{c}{\textbf{NIQE}$\downarrow$} \\ \midrule
\multicolumn{1}{l|}{\multirow{3}{*}{CycleGAN}} &
  IN\mbox{*} &
  \textbf{136.32} &
  0.26 &
  11.91 &
  \textbf{21.83} &
  13.64 &
  \textbf{142.28} &
  0.28 &
  10.57 &
  \textbf{23.73} &
  13.76 \\
\multicolumn{1}{l|}{} &
  TIN &
  220.00 &
  \textbf{0.28} &
  \textbf{5.42} &
  39.05 &
  12.01 &
  207.03 &
  \textbf{0.38} &
  \textbf{4.65} &
  41.16 &
  11.99 \\
\multicolumn{1}{l|}{} &
  KIN &
  157.26 &
  0.14 &
  7.22 &
  27.89 &
  \textbf{\underline{11.27}} &
  150.93 &
  0.19 &
  6.31 &
  29.87 &
  \textbf{\underline{11.54}} \\ \midrule
\multicolumn{1}{l|}{\multirow{3}{*}{CUT}} &
  IN\mbox{*} &
  \textbf{105.22} &
  \textbf{0.85} &
  14.81 &
  \textbf{23.76} &
  13.99 &
  105.66 &
  \textbf{0.85} &
  13.48 &
  \textbf{24.02} &
  14.05 \\
\multicolumn{1}{l|}{} &
  TIN &
  214.22 &
  0.54 &
  \textbf{10.02} &
  34.52 &
  \textbf{13.37} &
  200.56 &
  0.64 &
  \textbf{8.64} &
  35.01 &
  \textbf{13.37} \\
\multicolumn{1}{l|}{} &
  KIN &
  108.20 &
  0.81 &
  12.84 &
  31.26 &
  13.70 &
  \textbf{\underline{100.90}} &
  0.80 &
  11.58 &
  31.94 &
  13.86 \\ \midrule
\multicolumn{1}{l|}{\multirow{3}{*}{L-LSeSim}} &
  IN\mbox{*} &
  \textbf{107.74} &
  0.41 &
  11.83 &
  \textbf{21.09} &
  10.70 &
  \textbf{105.59} &
  \textbf{0.48} &
  10.67 &
  \textbf{21.25} &
  10.62 \\
\multicolumn{1}{l|}{} &
  TIN &
  203.70 &
  0.10 &
  \textbf{8.22} &
  24.87 &
  10.94 &
  191.44 &
  0.19 &
  \textbf{7.58} &
  23.34 &
  10.75 \\
\multicolumn{1}{l|}{} &
  KIN &
  113.92 &
  \textbf{\underline{0.41}} &
  8.64 &
  26.00 &
  \textbf{\underline{10.63}} &
  106.90 &
  0.46 &
  7.69 &
  26.85 &
  \textbf{\underline{10.40}} \\ \bottomrule
\end{tabular}%
\begin{tablenotes}
  \footnotesize
  \item IN\mbox{*}: Patch-wise IN; Corr.: Histogram correlation; Grad.: Sobel gradients; $\downarrow$: the lower the better; $\uparrow$: the higher the better.
\end{tablenotes}
\end{threeparttable}
}
\end{table}

\begin{table}[!htb]
\centering
\caption{\textbf{Quantitative results for Kyoto summer2autumn dataset.}  For each experiment, the bold shows the best performance; the underline indicates that KIN surpasses IN and TIN.}
\label{tab:kyoto}
\resizebox{0.78\textwidth}{!}{%
\begin{threeparttable}
\begin{tabular}{llrrrrrrrr}
\toprule
& 
  \textbf{} & \multicolumn{4}{c}{\textbf{Kyoto (training set)}} & 
  \multicolumn{4}{c}{\textbf{Kyoto (testing set)}} \\ 
  \cmidrule(l){3-6} \cmidrule(l){7-10} 
&
  \textbf{} &
  \multicolumn{1}{c}{\textbf{FID}$\downarrow$} &
  \multicolumn{1}{c}{\textbf{Grad.}$\downarrow$} &
  \multicolumn{1}{c}{\textbf{PIQE}$\downarrow$} &
  \multicolumn{1}{c}{\textbf{NIQE}$\downarrow$} &
  \multicolumn{1}{c}{\textbf{FID}$\downarrow$} &
  \multicolumn{1}{c}{\textbf{Grad.}$\downarrow$} &
  \multicolumn{1}{c}{\textbf{PIQE}$\downarrow$} &
  \multicolumn{1}{c}{\textbf{NIQE}$\downarrow$} \\ \midrule
\multicolumn{1}{l|}{\multirow{3}{*}{CycleGAN}} & 
  IN\mbox{*} &
  \textbf{79.11} &
  18.40 &
  \textbf{43.62} &
  12.20 &
  \textbf{171.88} &
  16.52 &
  \textbf{37.00} &
  11.26 \\
\multicolumn{1}{l|}{} &
  TIN &
  87.10 &
  \textbf{12.39} &
  54.29 & 
  12.24 & 
  180.50 & 
  \textbf{10.49} & 
  52.56 &
  11.92 \\
\multicolumn{1}{l|}{} &
  KIN &
  93.60 &
  17.08 &
  44.65 &
  \textbf{\underline{12.11}} &
  192.25 &
  15.12 &
  39.53 &
  \textbf{\underline{11.10}} \\
  \midrule
\multicolumn{1}{l|}{\multirow{3}{*}{CUT}} & 
  IN\mbox{*} &
  77.59 &
  17.87 & 
  43.81 &
  13.40 &
  \textbf{157.04} &
  18.29 &
  \textbf{37.74} &
  \textbf{11.41} \\
\multicolumn{1}{l|}{} &
  TIN &
  98.37 &
  \textbf{17.44} &
  43.30 &
  \textbf{12.11} &
  181.13 &
  \textbf{15.33} &
  40.97 &
  11.89 \\
\multicolumn{1}{l|}{} &
  KIN & 
  \textbf{\underline{75.27}} &
  18.53 &
  \textbf{\underline{40.21}} &
  12.98 &
  167.15 &
  17.24 &
  38.30 & 
  12.31 \\
  \midrule
\multicolumn{1}{l|}{\multirow{3}{*}{L-LSeSim}} &
  IN\mbox{*} & 
  \textbf{178.19} & 
  14.86 & 
  19.35 & 
  \textbf{11.89} & 
  \textbf{248.81} & 
  13.05 & 
  14.14 & 
  11.42 \\
\multicolumn{1}{l|}{} &
  TIN &
  178.98 &
  \textbf{11.27} &
  \textbf{19.13} & 
  12.07 &
  253.14 &
  \textbf{9.74} &
  \textbf{12.21} & 
  11.18 \\
\multicolumn{1}{l|}{} & 
  KIN &
  192.42 & 
  16.41 & 
  19.19 & 
  12.01 & 
  265.07 & 
  16.12 & 
  13.00 &
\textbf{\underline{10.54}} \\
\bottomrule
\end{tabular}%
\begin{tablenotes}
  \footnotesize
  \item IN\mbox{*}: Patch-wise IN; Grad.: Sobel gradients; $\downarrow$: the lower the better
\end{tablenotes}
\end{threeparttable}
}
\end{table}

\begin{figure}[!htb]
  \centering
  \subfigure[Quality evaluation]{\label{fig:quality_evaluation}\includegraphics[width=0.58\textwidth]{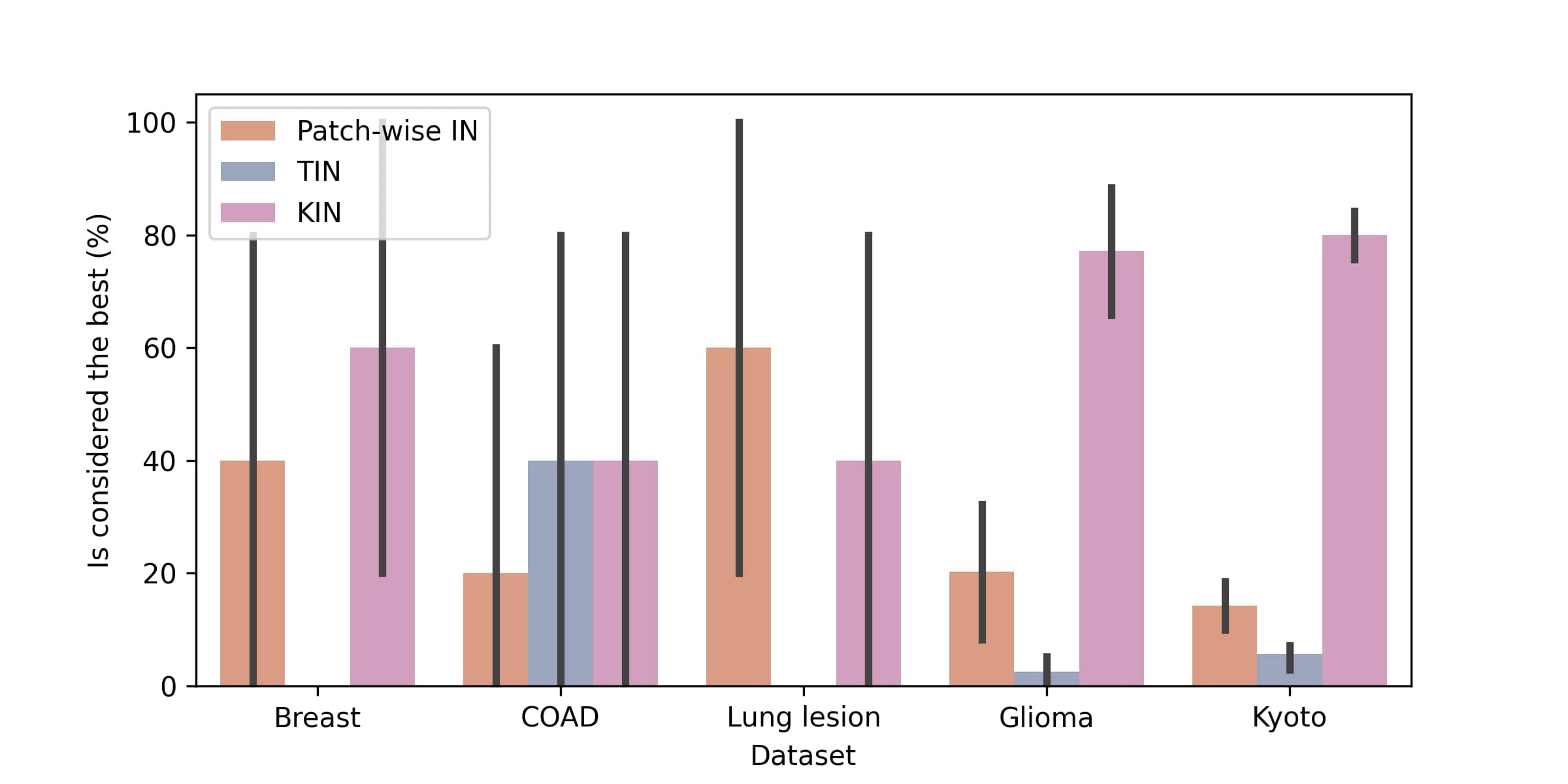}}                
  \subfigure[Fidelity evaluation]{\label{fig:fidelity_evaluation}\includegraphics[width=0.36\textwidth]{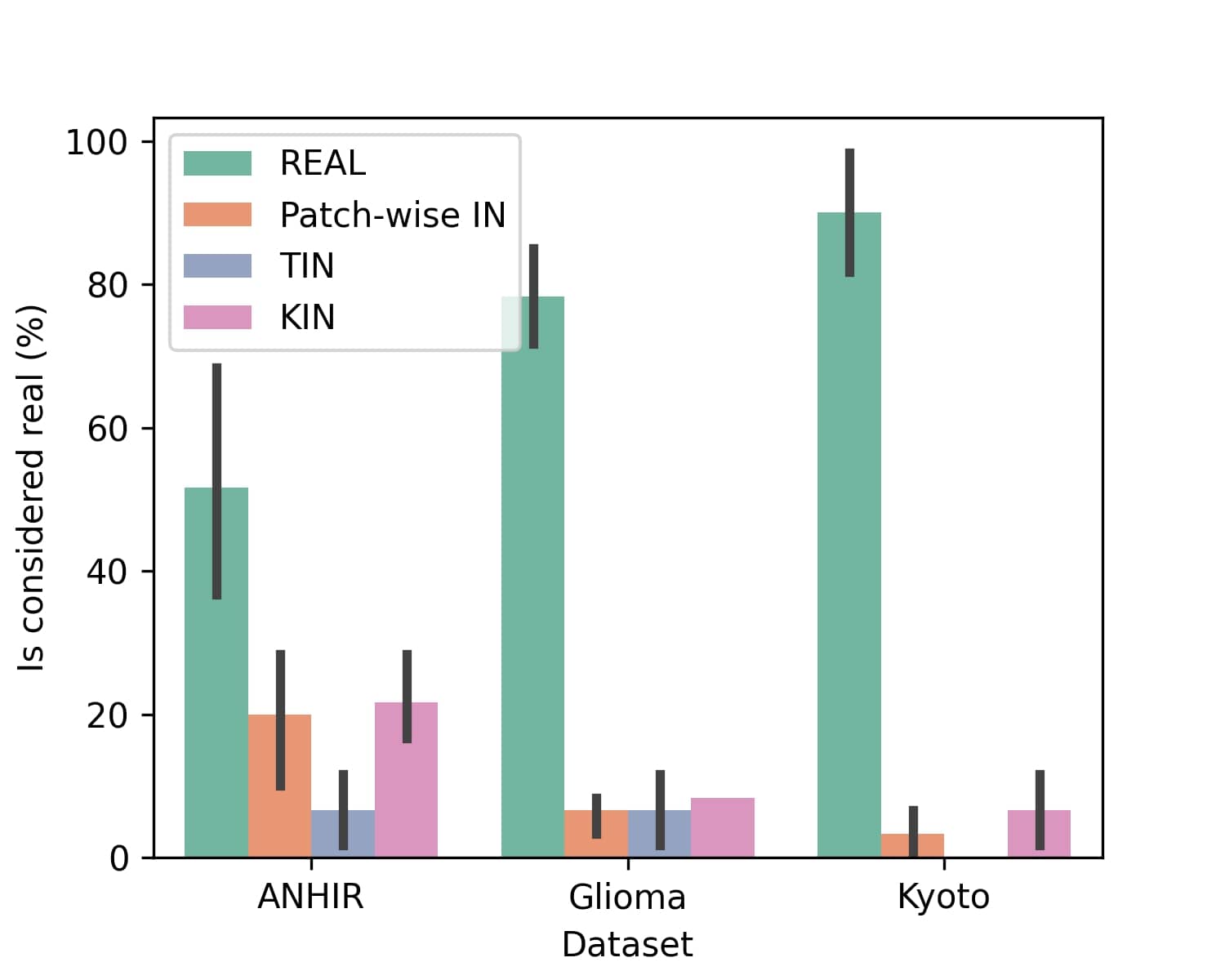}}
  \caption{\textbf{Human evaluation results.} In quality evaluation, KIN achieved the best or competitive performance among all datasets while images generated via TIN obtained the worst quality. For the fidelity evaluation, although real consecutive section of tissue is easily to be distinguished from the fake ones, KIN is still the most deceptive among all methods. It can be noticed that translated natural images are hardly to deceive human since their complicated content are difficult to be fabricated.}
  \label{fig:user_study}
\end{figure}

\subsubsection{Stain transformation}
\Cref{fig:breast,fig:COAD,fig:lung_lesion} and Fig. S5 and Fig. S6 show the translated images for three ANHIR subdatasets (breast tissue, COAD, and lung lesion) and glioma dataset, respectively.
With only IN layers, CUT yields the images with best quality with some tiling artifacts, while CycleGAN led to checkerboard artifacts. L-LSeSim powerfully preserves spatial information but compromises color information.
With TIN, all the translated images showed dramatically over/under-colorizing.
With our KIN, translated images can have minor tiling artifacts and preserve their local features.
However, if the original framework generated severe tiling artifacts, our KIN could alleviate but be hard to eliminate.
Considering the similarity (FID and histogram correlation) and quality metrics (Sobel gradient, PIQE and NIQE), our KIN is superior to patch-wise IN and TIN in most cases (see \Cref{tab:ANHIR,tab:glioma}).
Although KIN does not always obtain the best scores, a possible reason is that no appropriate metrics can reflect the performance of such unpaired WSIs stain transformation task.
Thus, we established two human evaluation studies to pertinently evaluate the image quality and fidelity. As shown in Fig.~\ref{fig:user_study}, our KIN achieved the best performance in both.

\subsubsection{Translation for natural images}
Our KIN module also performed well on natural images (as shown in Fig.~\ref{fig:kyoto_test} and Fig. S7). As described above, KIN can alleviate the tiling artifacts generated by patch-wise IN while TIN would lead to over/under-colorizing.
However, when it comes to natural images, over/under-colorizing would not be as obtrusive as in stain transformation cases, since people sometimes prefer over-stylized images. For example, High-Dynamic Range (HDR) or contrast adjustment techniques are popular to beautify photographs and render the photos more attractive.
Tab.~\ref{tab:kyoto} and Fig.~\ref{fig:user_study} provided the metrics evaluation results, and our KIN obtained the best performance among all methodologies in human evaluation.
Considering the FID, CUT with our KIN is superior or competitive to other methods. Although Sobel gradients are higher in some cases, the high contrast level of one image might also contribute to higher gradients. On the other hand, there are only minor differences in PIQE and NIQE between methods. However, none of the metrics can effectively evaluate ultra-high-resolution images with tiling artifacts.

\subsection{Ablation study}

\subsubsection{Kernel and kernel size}
To elucidate the effect of different kernels and kernel size on the translated images, we applied constant and Gaussian kernels with the size of 1, 3, 7, 11, and $\infty$ in the KIN module (see Fig. S8 and S9). It is noteworthy that when kernel size is set to 1, the KIN module will operate in a manner of patch-wise IN, whereas it would be like TIN when kernel size is set to $\infty$ (bounded by the input image size).
KIN is an eclectic approach that combines the advantages of patch-wise IN and TIN and avoids extremes of single and global features calculated in patch-wise IN and TIN. When kernel size increases from one to $\infty$, the translated results gradually change from patch-wise IN to TIN. On the other hand, the constant kernel can help generate smoother results, while the Gaussian kernel will emphasize local features more.

\section{Discussions}
%

%
Our experiments showed that KIN performed well on multiple datasets, and unseen testing data can even be successfully inferred when sufficient training data are available. The over/under-colorizing problem caused by TIN is also revealed, which might be innocuous when natural images are used but would be detrimental when targeting stain transformation. Pathological features, which are essential for clinical judgment, would be compromised when global mean and standard deviation are applied in the TIN. 
Although KIN can be inserted into any IN-based framework, the performance would be compromised if the original framework has amateurish performance, such as CycleGAN, which generates results diversely among adjacent patches. KIN can hardly eliminate all the tiling artifacts undertaking such cases.
Interestingly, we found that CUT can yield more consistent results among adjacent patches, especially for the hue. On the other hand, LSeSim meticulously preserves all the structure but ignores the consistency of the hue, which is reasonable as CUT captures domain-specific features, but LSeSim focuses on spatial features according to their loss functions.
Despite KIN achieving the best performance surpassing all previous approaches in human evaluation studies, its strength cannot be manifested due to the inadequacy of appropriate metrics for evaluating the quality and fidelity of unpaired ultra-high-resolution WSIs.
Finally, ultra-high-resolution images are commonly used in daily life but there is no public dataset available for a fair comparison. To facilitate related researches, we released the Kyoto summer2autumn dataset.

\section{Conclusion}
This study presents Kernelized Instance Normalization (KIN) for ultra-high-resolution stain transformation with constant space complexity. KIN can be easily inserted into popular unpaired image-to-image translation frameworks without re-training the model. Comprehensive experiments with two WSI datasets were conducted and evaluated by human evaluation studies and appropriate metrics. An extra ultra-high-resolution natural image dataset was also utilized and demonstrated the generalizability of KIN. Overall, KIN surpassed all the previous approaches and generated state-of-the-art outcomes. Henceforth, ultra-high-resolution stain transformation or image-to-image translation, can be easily accomplished and applied in clinical practice.
\paragraph{\textbf{Acknowledgements}}
We thank Chao-Yuan Yeh, the CEO of aetherAI, for providing computing resources, which enabled this study to be performed, and Cheng-Kun Yang for his revision suggestions.

\clearpage
%
%
\bibliographystyle{splncs04}
\bibliography{eccv2022submission}

\end{document}


\pagestyle{headings}
\mainmatter
\def\ECCVSubNumber{2310}  

\title{Ultra-high-resolution unpaired stain transformation via Kernelized Instance Normalization (Supplementary Material)
} 

\titlerunning{Ultra-high-resolution unpaired stain transformation}
%
\author{Ming-Yang Ho\thanks{%
    Corresponding author} \and
Min-Sheng Wu \and
Che-Ming Wu}
\authorrunning{Ho. et al.}
%
\institute{aetherAI, Taipei, Taiwan\\
\email{\{kaminyouho,vincentwu,uno\}@aetherai.com}\\
\url{https://www.aetherai.com/}}
\maketitle

\subsection{Analysis}

\begin{figure}[!htb]
  \centering
  \subfigure[Mean (layer 1)]{\label{fig:block1_mean}\includegraphics[width=0.49\textwidth]{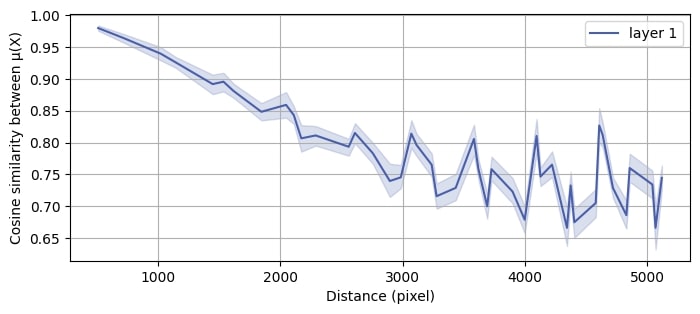}}
  \subfigure[Std (layer 1)]{\label{fig:block1_std}\includegraphics[width=0.49\textwidth]{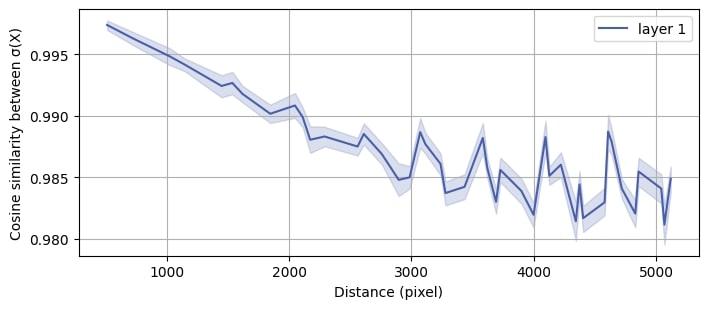}}
  \subfigure[Mean (layer 2-6)]{\label{fig:blocks_mean}\includegraphics[width=0.49\textwidth]{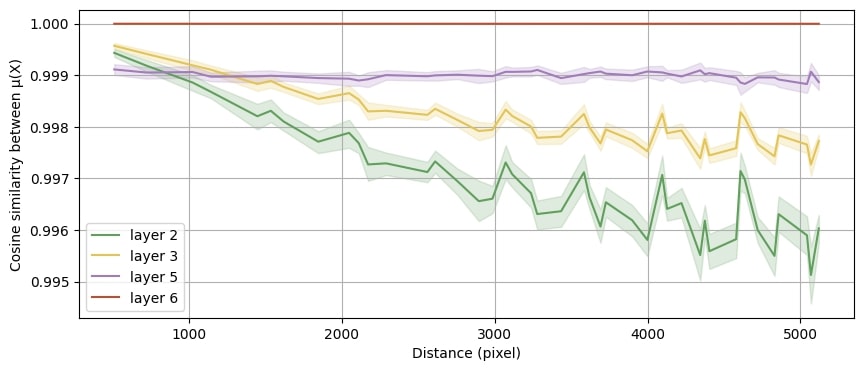}}
  \subfigure[Std (layer 2-6)]{\label{fig:blocks_std}\includegraphics[width=0.49\textwidth]{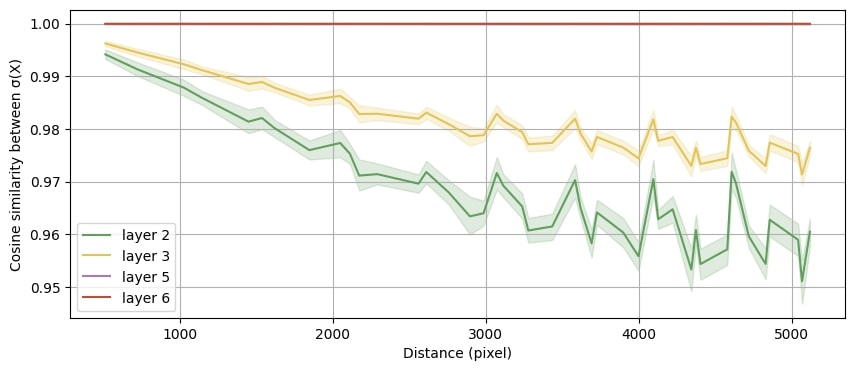}}
  \caption{\textbf{Comparison of mean and std calculated in IN between adjacent patches.} Mean, and standard deviation (std) of every two adjacent or nearby patches (up to 5,000 pixels far away) were extracted from the IN in the original CycleGAN model and compared. The CycleGAN model comprises 6 layers and each has one or multiple IN: (1) convolutional layer; (2) down-sampling layer; (3) down-sampling layer; (4) residual backbone; (5) up-sampling layer; (6) up-sampling layer. We analyzed mean and std from IN in all layers except the fourth layer, which is a backbone. It can be noticed that there is a great discrepancy in mean and std between faraway patches in the earlier layers.}
  \label{fig:cosine_similarity_patches}
\end{figure}

\begin{figure}[!htb]
\centering
\resizebox{0.5\textwidth}{!}{%
\includegraphics[]{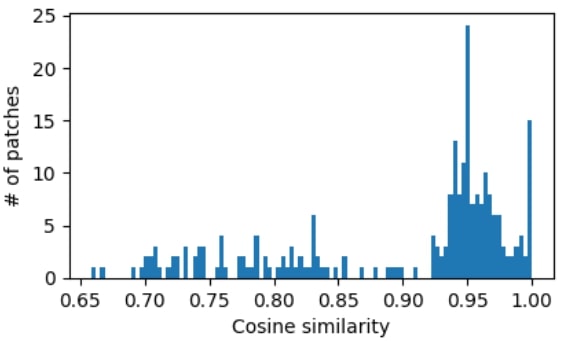}
}
\caption{\textbf{Distribution of cosine similarity between means of thumbnail and patches calculated in IN.} The means of patches and the thumbnail calculated in the IN layer from layer 1 of CycleGAN's generator are extracted and compared. Distribution of the cosine similarity is shown. An obvious discrepancy can be observed, which indicates the inappropriateness of using thumbnail statistics for all cropped patches in the TIN [8].}
\label{fig:cosine_similarity_thumbnail}
\end{figure}

To verify our hypothesis, we extracted the $\mu(X)$ and $\sigma(X)$ from all the IN layers in $\mathcal{G}$ for patches cropped from one single image, in which $\mu(X), \sigma(X) \in R^{1 \times C}$, and $C$ is the number of channels. Then, $\mu(X)$ and $\sigma(X)$ were further flattened into vectors with size $C$ to compute the cosine similarity between every pair. Besides, the Euclidean distances between pairs were recorded.
Fig.~\ref{fig:cosine_similarity_patches} demonstrates that cosine similarity of $\mu(X)$ and $\sigma(X)$ between two patches would dramatically decrease when two patches are farther apart, especially in the first few layer blocks.

In addition, we adopted the methodology proposed in TIN [8] and measured the $\mu(X)$ and $\sigma(X)$ between the thumbnail and other cropped patches in Figure \ref{fig:cosine_similarity_thumbnail}.
It shows that extreme inconsistency occurs in the first few layers, implying local contrast and hue information will diminish if $\mu$ and $\sigma$ of thumbnail are used.
On the contrary, the convolution mechanism in our KIN can both alleviate this inconsistency issue and further improve the assembly quality when adjacent patches are combined.

\subsection{Performance on the classification downstream task}
As there is no well-developed metric that can evaluate unpaired ultra-high-resolution (UHR) images, downstream classification task was experimented to address this issue. We conducted a classification task for the ANHIR dataset (breast, lung lesion, and COAD). A ResNet-50 model was trained on the patches cropped from real WSIs in the IHC domain and tested on the patches cropped from translated WSIs generated by patch-wise IN, TIN, and KIN with the CUT framework. We deliberately cropped patches from the attached boundary to evaluate the influence of tilting artifacts.
The accuracies of patch-wise IN, TIN, and \textbf{KIN} are 98.8\%, 88.4\%, and \textbf{99.2}\%, respectively. The results show that KIN achieves the best performance, which might be due to the reduction of tilting artifacts that confused the classifier. TIN obtains the worst performance since using global statistics might lead to the loss of local information.

\subsection{Evaluated by SSIM and FSIM metrics}
\begin{figure}[!htb]
  \centering
  \subfigure[Patch-wise IN]{\label{fig:gray2rgb_in}\includegraphics[width=0.30\textwidth]{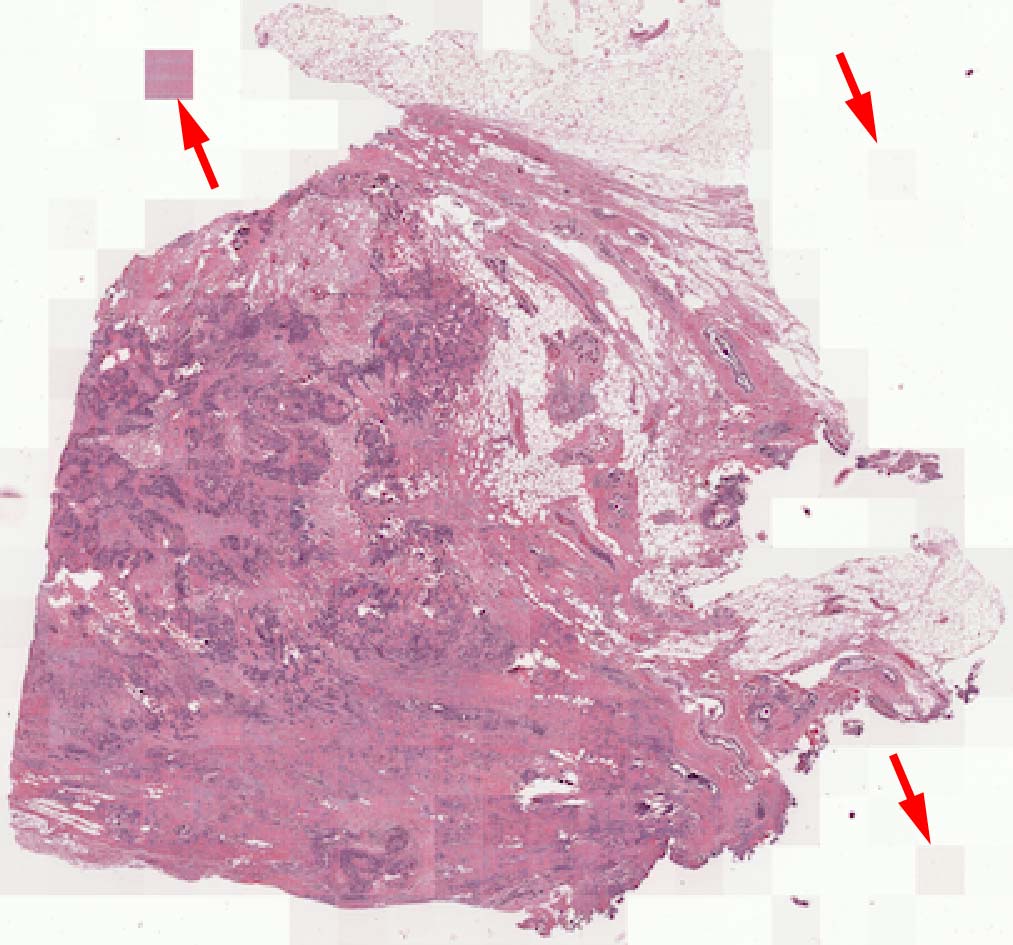}}
  \subfigure[TIN]{\label{fig:gray2rgb_tin}\includegraphics[width=0.30\textwidth]{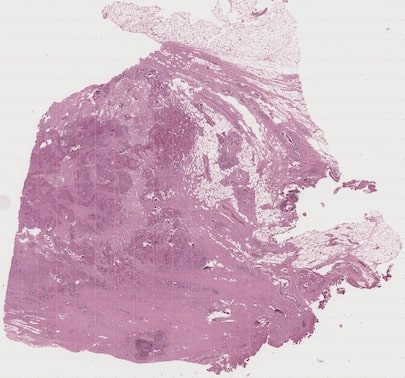}}
  \subfigure[KIN]{\label{fig:gray2rgb_kin}\includegraphics[width=0.30\textwidth]{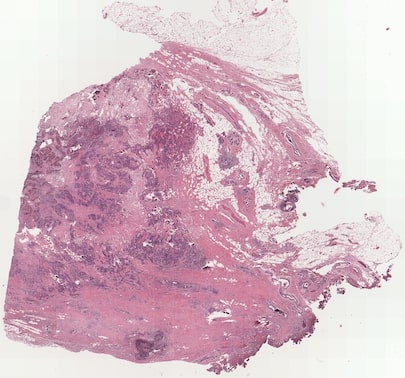}}
  \caption{\textbf{Generated RGB WSIs by different methods.} The presence of tilting artifacts, indicated by red arrows,  cannot evaluated by SSIM or FSIM metrics.}
  \label{fig:gray2rgb}
\end{figure}

To evaluate KIN with SSIM and FSIM metrics, we experimented with pairwise translating gray images of the ANHIR dataset into H\&E. However, both SSIM (patch-wise IN: 0.94, TIN: 0.90, KIN: 0.93) and FSIM (patch-wise IN: 0.79, TIN: 0.74, KIN: 0.78) cannot evaluate the presence of tilting artifacts in patch-wise IN (see Fig. ~\ref{fig:gray2rgb}).

\subsection{Failure modes of KIN}
\begin{figure}[!htb]
  \centering
  \subfigure[Source]{\label{fig:failure_source}\includegraphics[width=0.48\textwidth]{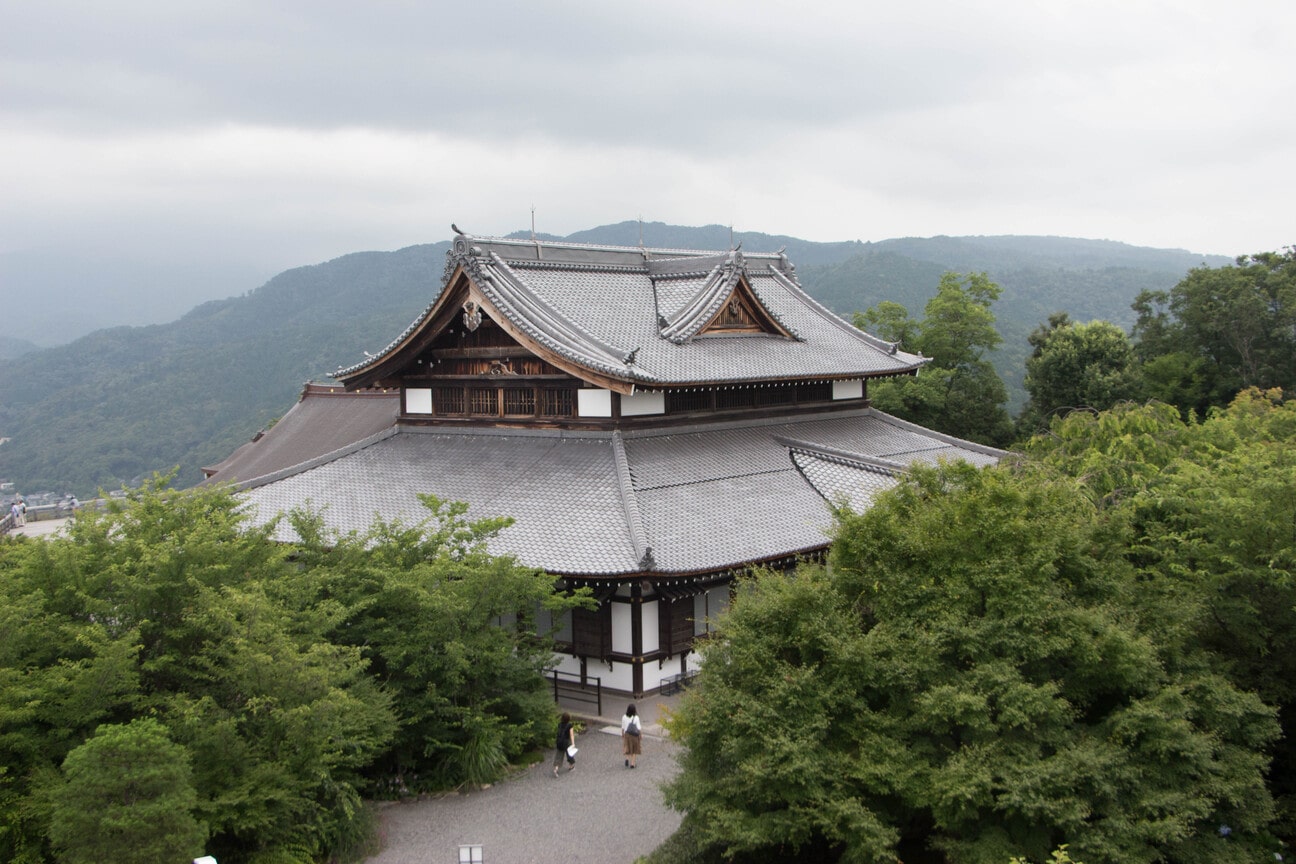}}
  \subfigure[Patch-wise IN]{\label{fig:failure_in}\includegraphics[width=0.48\textwidth]{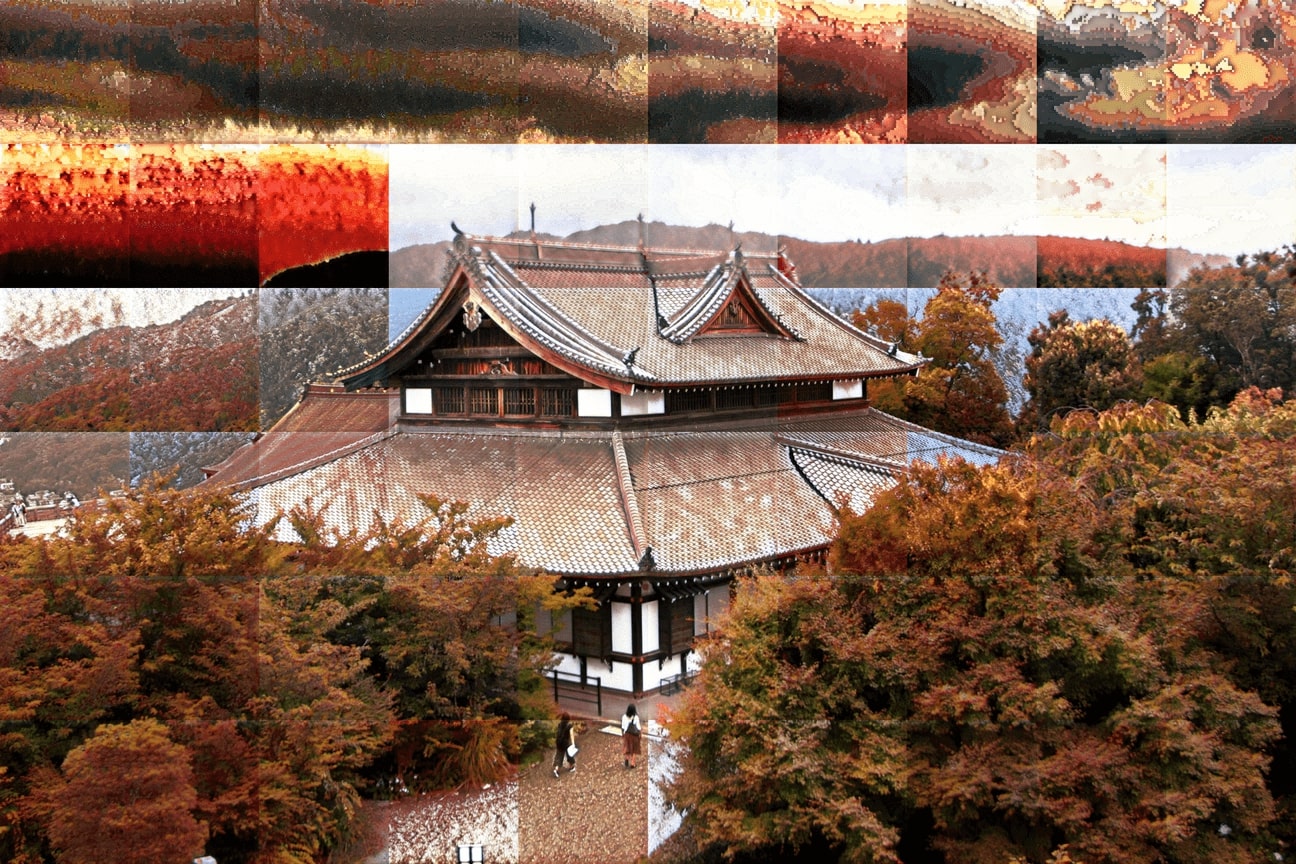}}
  \subfigure[TIN]{\label{fig:failure_tin}\includegraphics[width=0.48\textwidth]{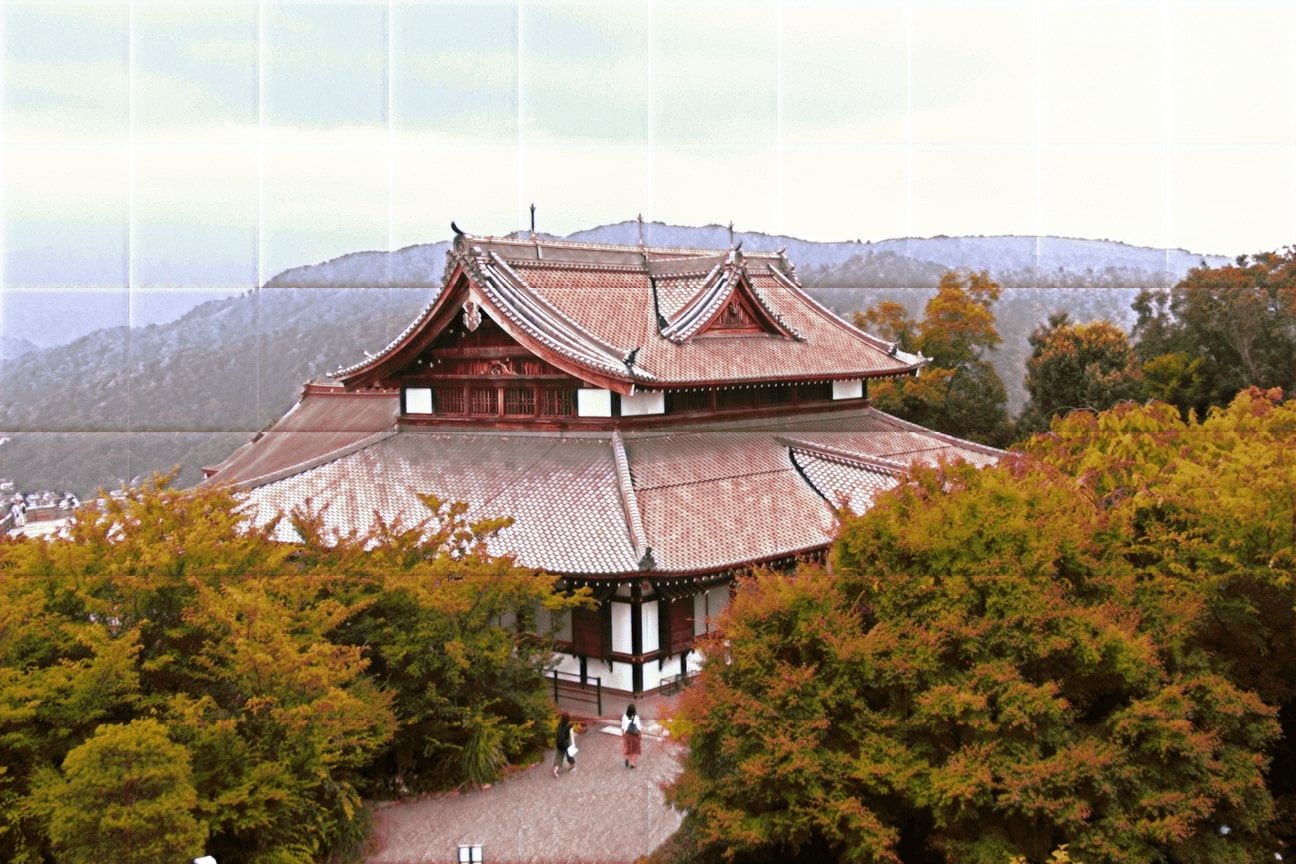}}
  \subfigure[KIN]{\label{fig:failure_kin}\includegraphics[width=0.48\textwidth]{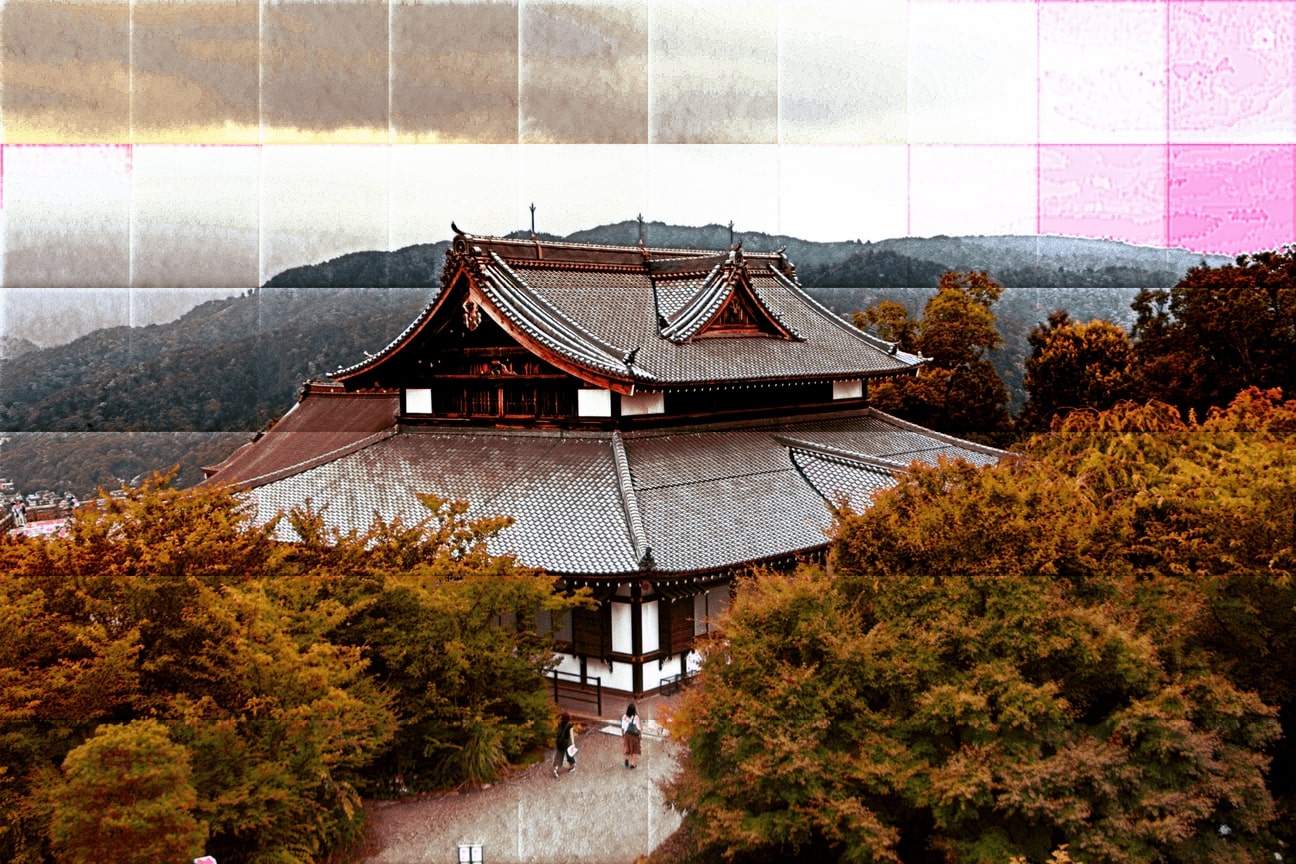}}
  \caption{\textbf{Failure modes.} KIN will be inferior to TIN if training data lack enough specific scene.}
  \label{fig:failure}
\end{figure}

If the training data lack enough specific scene (e.g., the sky in Kyoto dataset), KIN will be inferior to TIN (see Fig. ~\ref{fig:failure}).

\begin{figure}[!htb]
\centering
\resizebox{\textwidth}{!}{%
\includegraphics[]{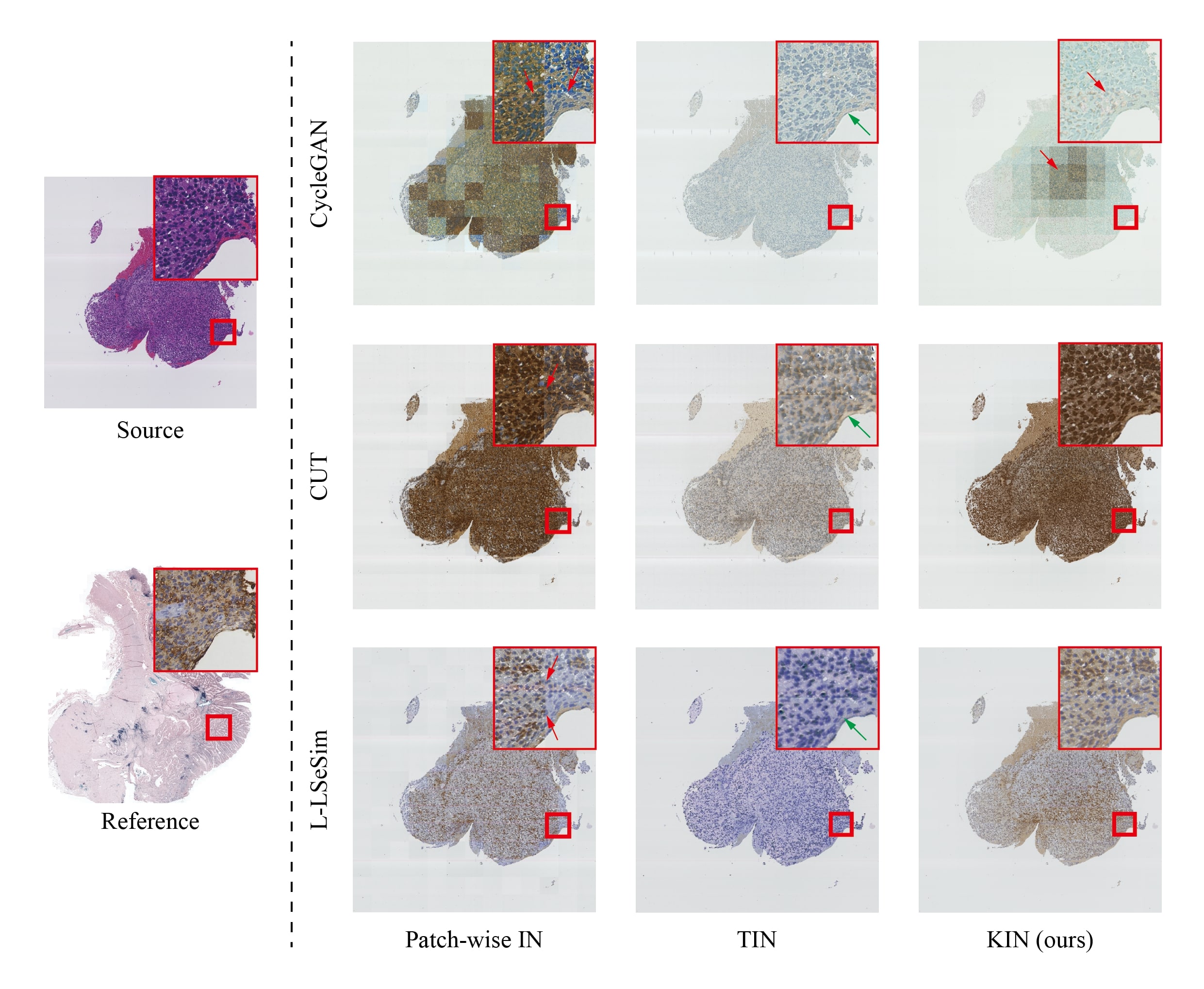}
}
\caption{\textbf{H\&E-to-EGFR stain transformation results on Glioma training set ($\mathbf{7,755 \times 7,109}$ pixels)} generated by different frameworks with IN, TIN, and KIN layers. Red arrows indicate tilting artifacts; green arrows indicate over/under-colorizing. CUT+KIN achieved the best performance. Zoom in for better view.}
\label{fig:glioma_train}
\end{figure}

\begin{figure}[!htb]
\centering
\resizebox{0.9\textwidth}{!}{%
\includegraphics[]{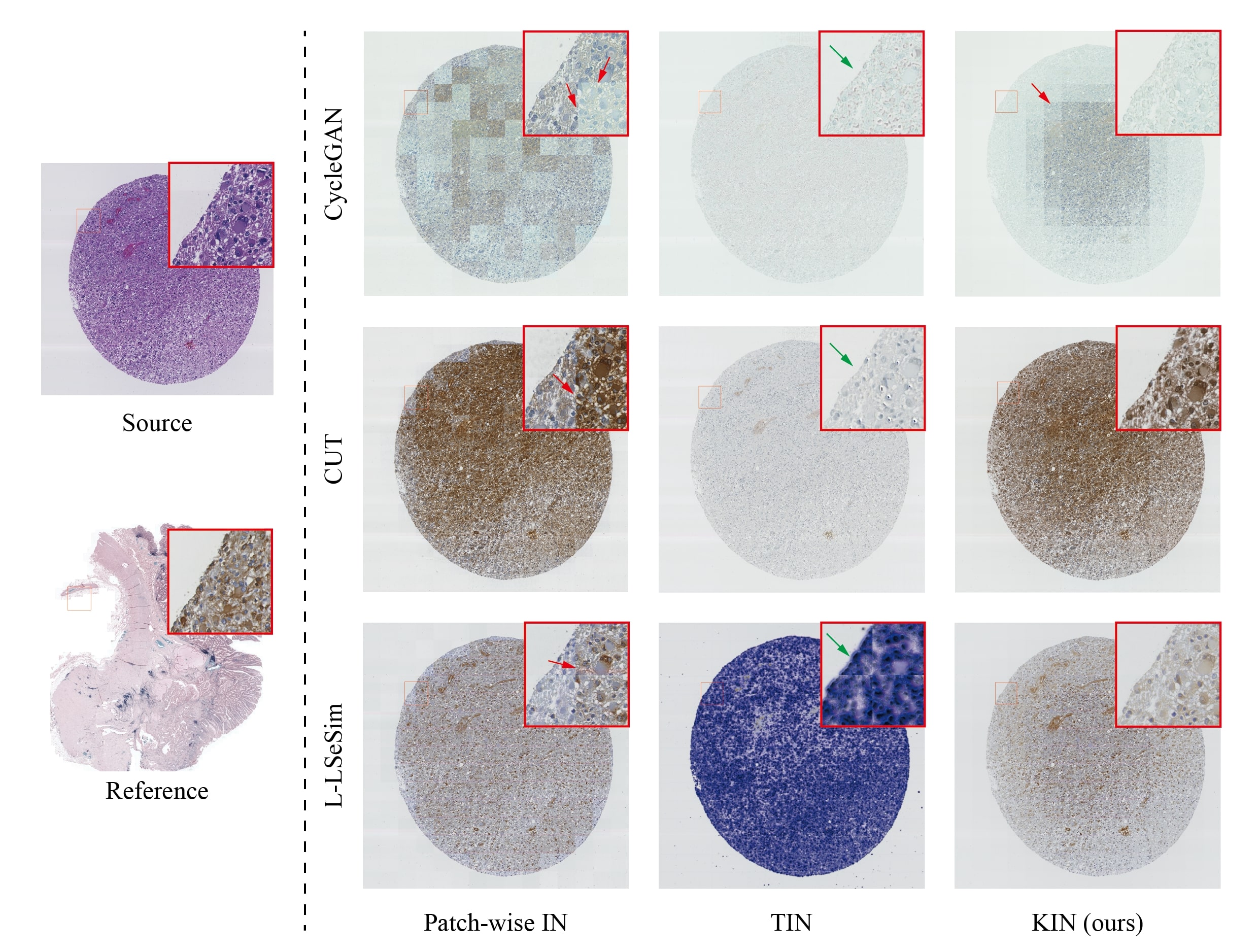}
}
\caption{\textbf{H\&E-to-EGFR stain transformation results on Glioma testing set ($\mathbf{8,078 \times 8,078}$ pixels)} generated by different frameworks with IN, TIN, and KIN layers. Red arrows indicate tiling artifacts; green arrows indicate over/under-colorizing. CUT+KIN achieved the best performance. Zoom in for better view.}
\label{fig:glioma_test}
\end{figure}

\begin{figure}[!htb]
\centering
\resizebox{\textwidth}{!}{%
\includegraphics[]{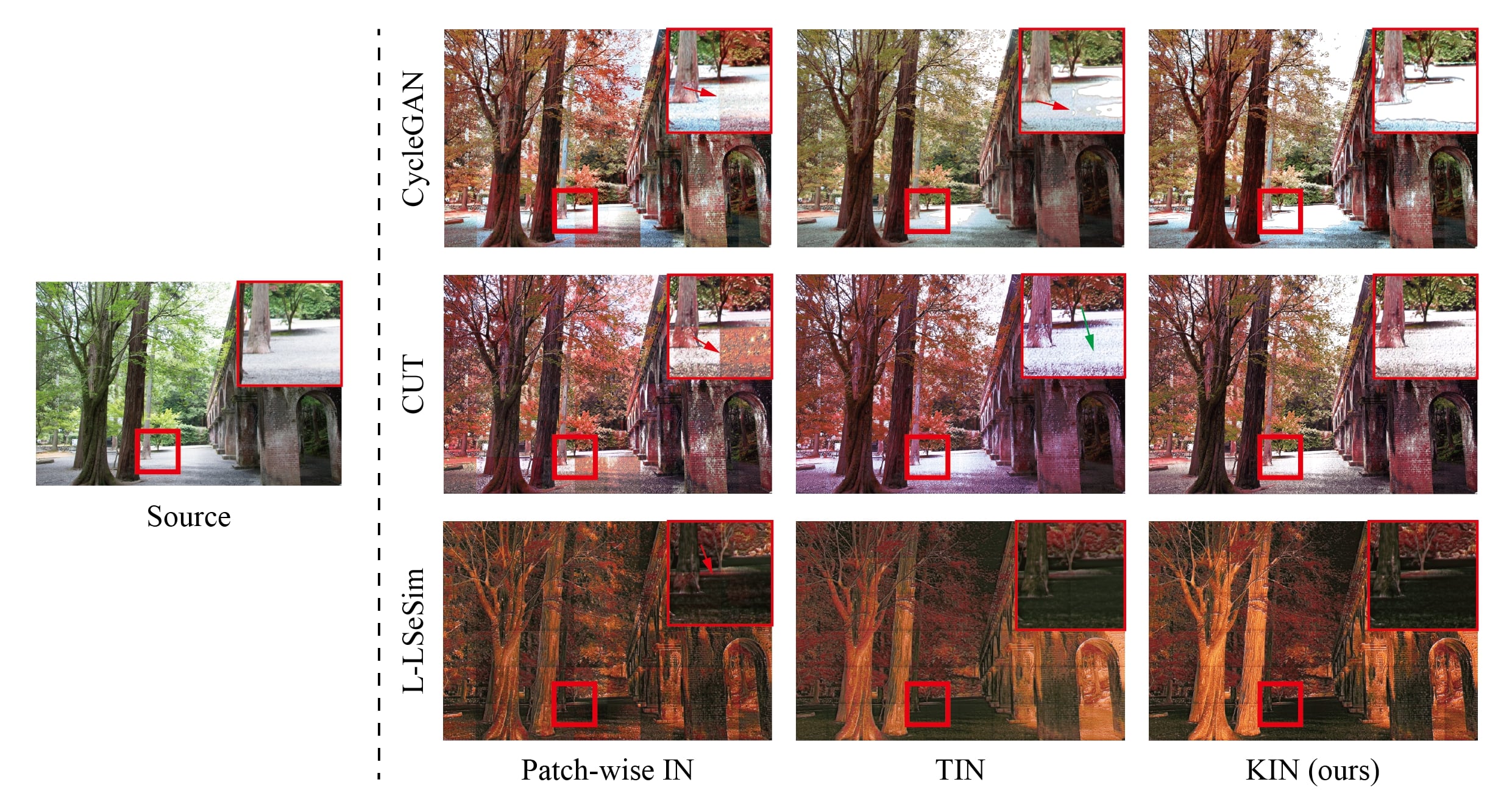}
}
\caption{\textbf{Image-to-image translation results on Kyoto summer2autumn training set ($\mathbf{3,456 \times 5,184}$ pixels)} generated by different frameworks with IN, TIN, and KIN layers. Red arrows indicate tilting artifacts; green arrows indicate over/under-colorizing. CUT+KIN achieved the best performance. Zoom in for better view.} 
\label{fig:kyoto_train}
\end{figure}

\begin{figure}[!htb]
\centering
\resizebox{\textwidth}{!}{%
\includegraphics[]{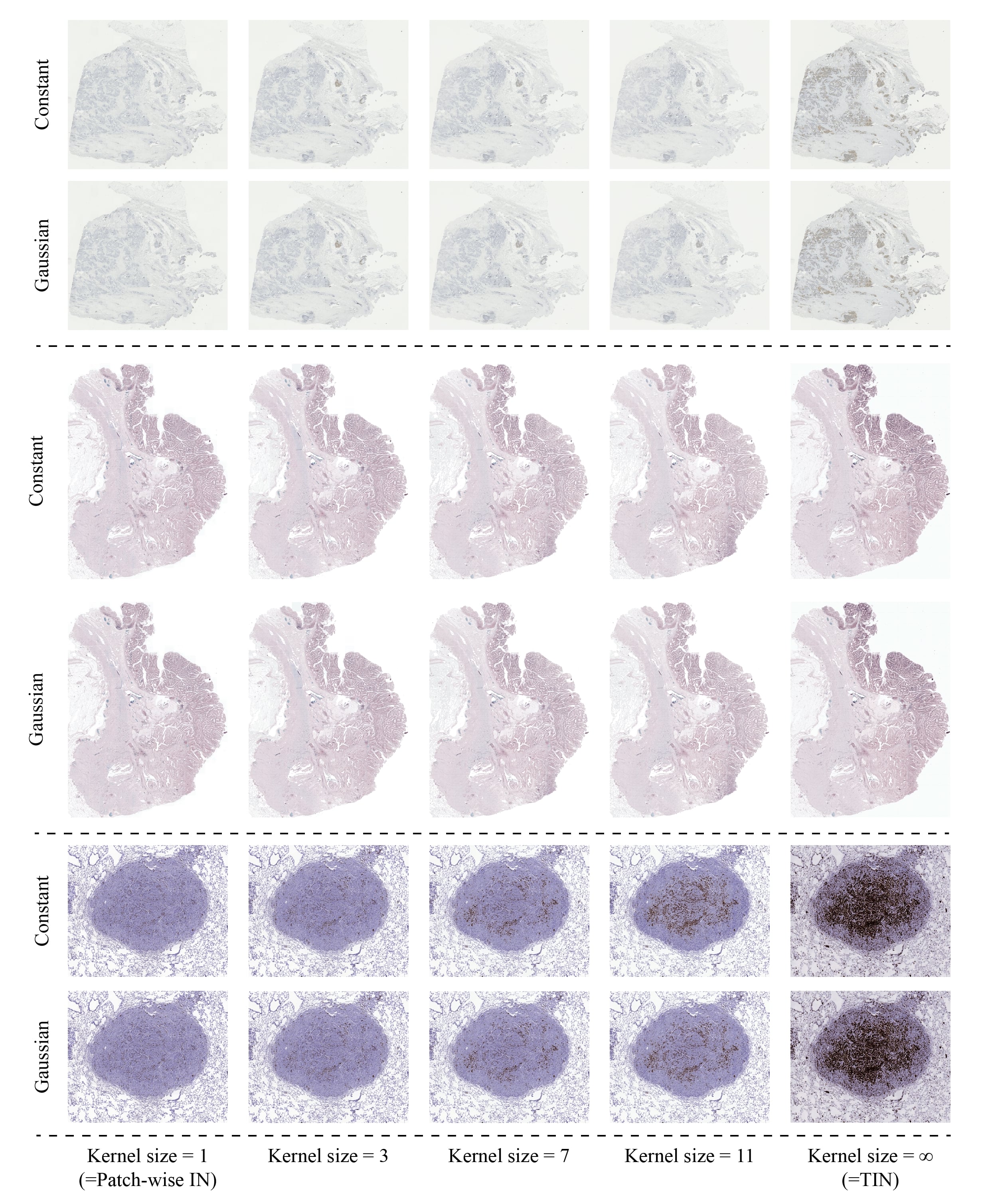}
}
\caption{\textbf{Ablation study for kernel types on three ANHIR subdatasets.} Constant and Gaussian kernels with the size of 1, 3, 7, 11, and $\infty$ are applied to elucidate the effect of KIN module. When kernel size is set to 1, the KIN module will operate in a manner of patch-wise IN, whereas it would be like TIN when kernel size is set to $\infty$.}
\label{fig:ablation_ANHIR}
\end{figure}

\begin{figure}[!htb]
\centering
\resizebox{\textwidth}{!}{%
\includegraphics[]{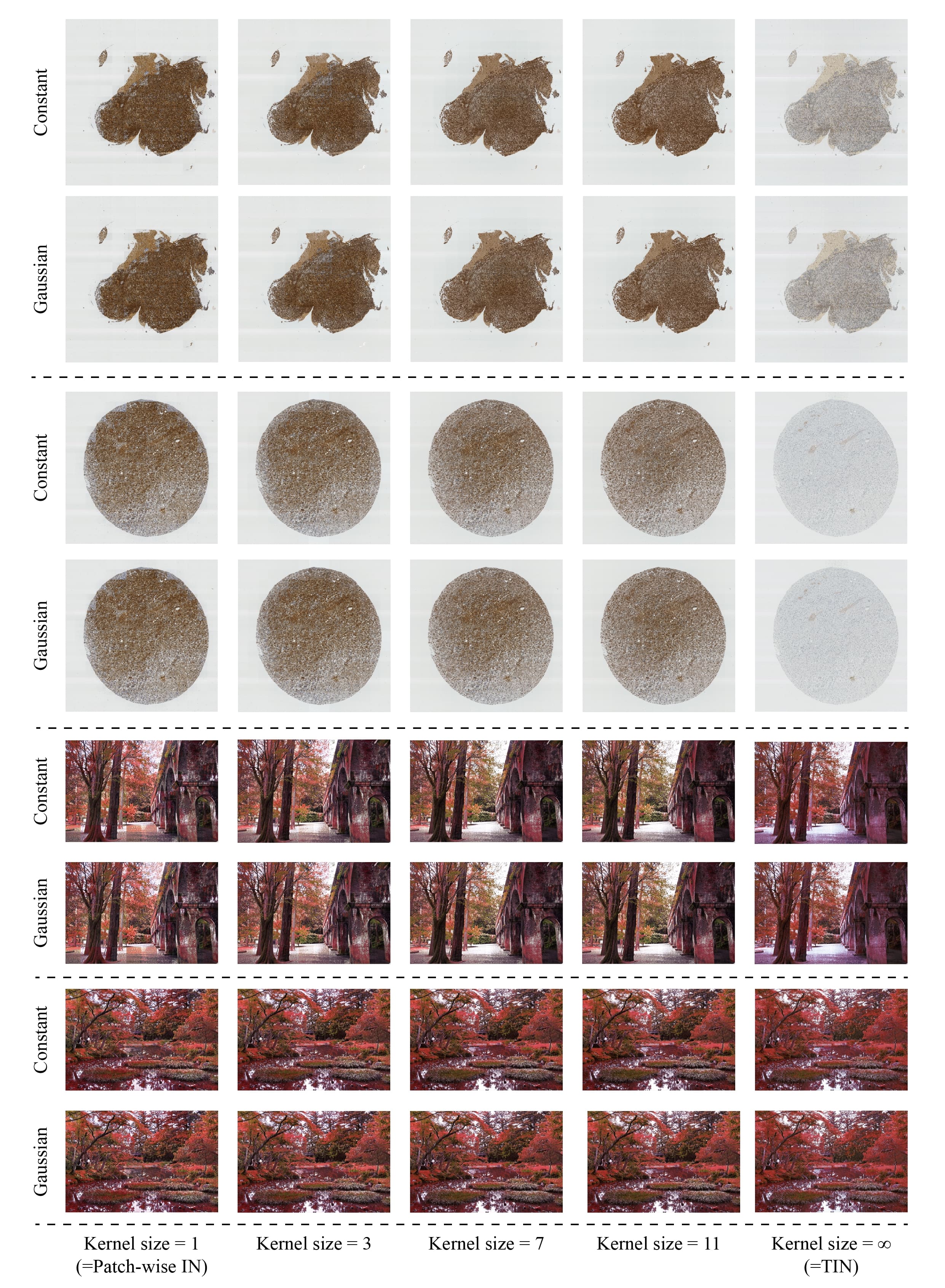}
}
\caption{\textbf{Ablation study for kernel types on Glioma and Kyoto summer2autumn datasets.} Constant and Gaussian kernels with the size of 1, 3, 7, 11, and $\infty$ are applied to elucidate the effect of KIN module. When kernel size is set to 1, the KIN module will operate in a manner of patch-wise IN, whereas it would be like TIN when kernel size is set to $\infty$.}
\label{fig:ablation_glioma_kyoto}
\end{figure}

%
%